\def\year{2022}\relax
\documentclass[letterpaper]{article} 
\usepackage{aaai22}  
\usepackage{times}  
\usepackage{helvet}  
\usepackage{courier}  
\usepackage[hyphens]{url}  
\usepackage{graphicx} 
\usepackage{natbib}  
\usepackage{caption} 
\DeclareCaptionStyle{ruled}{labelfont=normalfont,labelsep=colon,strut=off} 
\usepackage{algorithm}
\usepackage{algorithmic}

\usepackage{amsmath}
\usepackage{amsfonts}
\usepackage{amssymb}
\usepackage{multirow}
\usepackage{booktabs}
\usepackage{dsfont}
\usepackage{xcolor}

\usepackage{amsthm}
\usepackage{verbatim}

\DeclareMathOperator*{\argmax}{\arg\max}

\usepackage{newfloat}
\usepackage{listings}
\usepackage{wrapfig}
\floatstyle{ruled}
\newfloat{listing}{tb}{lst}{}
\floatname{listing}{Listing}
\title{Contrastive Regularization for Semi-Supervised Learning}

\author{
    \Large \textbf{Doyup Lee\textsuperscript{\rm 1}, 
                    Sungwoong Kim\textsuperscript{\rm 2}, 
                    Ildoo Kim\textsuperscript{\rm 2}, 
                    Yeongjae Cheon\textsuperscript{\rm 3}, 
                    Minsu Cho\textsuperscript{\rm 1}, 
                    Wook-Shin Han\textsuperscript{\rm 1}\thanks{Corresponding Author}} \\
}

\affiliations{
    POSTECH\textsuperscript{\rm 1}, Kakao Brain\textsuperscript{\rm 2}, DCML Inc.\textsuperscript{\rm 3} \\
    South Korea \\
    \{doyup.lee, mscho, wshan\}@postech.ac.kr\textsuperscript{\rm 1}, \{swkim, ildoo.kim\}@kakaobrain.com\textsuperscript{\rm 2}, yj.star@dcml.co.kr\textsuperscript{\rm 3}\\
}

\usepackage{bibentry}

\begin{document}
\maketitle

\begin{abstract}
  Consistency regularization on label predictions becomes a fundamental technique in semi-supervised learning, but it still requires a large number of training iterations for high performance.
  In this study, we analyze that the consistency regularization restricts the propagation of labeling information due to the exclusion of samples with unconfident pseudo-labels in the model updates.
  Then, we propose \textit{contrastive regularization} to improve both efficiency and accuracy of the consistency regularization by well-clustered features of unlabeled data.
  In specific, after strongly augmented samples are assigned to clusters by their pseudo-labels, our contrastive regularization updates the model so that the features with confident pseudo-labels aggregate the features in the same cluster, while pushing away features in different clusters.
  As a result, the information of confident pseudo-labels can be effectively propagated into more unlabeled samples during training by the well-clustered features.
  On benchmarks of semi-supervised learning tasks, our contrastive regularization improves the previous consistency-based methods and achieves state-of-the-art results, especially with fewer training iterations.
  Our method also shows robust performance on open-set semi-supervised learning where unlabeled data includes out-of-distribution samples.
  
\end{abstract}

\section{Introduction}
\label{sec:introduction}

Recent semi-supervised learning (SSL) methods mostly make use of the consistency regularization to learn a specific task with sparse labels, showing competitive results to the fully supervised learning \cite{berthelot2019remixmatch,sohn2020fixmatch,li2020comatch}.
The consistency regularization enforces a model to produce consistent predictions on various augmented views of input with pseudo-labeling \cite{lee2013pseudo}. 
Moreover, in order to avoid a confirmation bias \cite{arazo2020pseudo} and increase the reliability of pseudo-labeling, a selection mask is typically used in this consistency regularization to exclude unconfident label predictions during SSL training. 
Consequently, the consistency regularization can propagate the labeling information into unlabeled samples around the augmented views of confident pseudo-labels \cite{ghosh2021data}.

Despite its promising results, the existing consistency regularization requires an expensive training cost to achieve high performance.
For example, although FixMatch \cite{sohn2020fixmatch} can achieve high SSL performance without pretraining on a large scale unlabeled data \cite{chen2020big}, it needs over 10,000 epochs to obtain the best performance even on small-scale datasets such as SVHN, CIFAR-10, or CIFAR-100.
Thus, we first analyze the inefficiency of the consistency regularization for SSL, both theoretically and empirically, and then verify that this inefficiency is originated from the exclusion of samples with unconfident pseudo-labels when updating a model.
Namely, it restricts the active propagation of confident labeling information into unlabeled samples, especially in the early stage of training.

\begin{figure}
\centering
\includegraphics[width=3.3in]{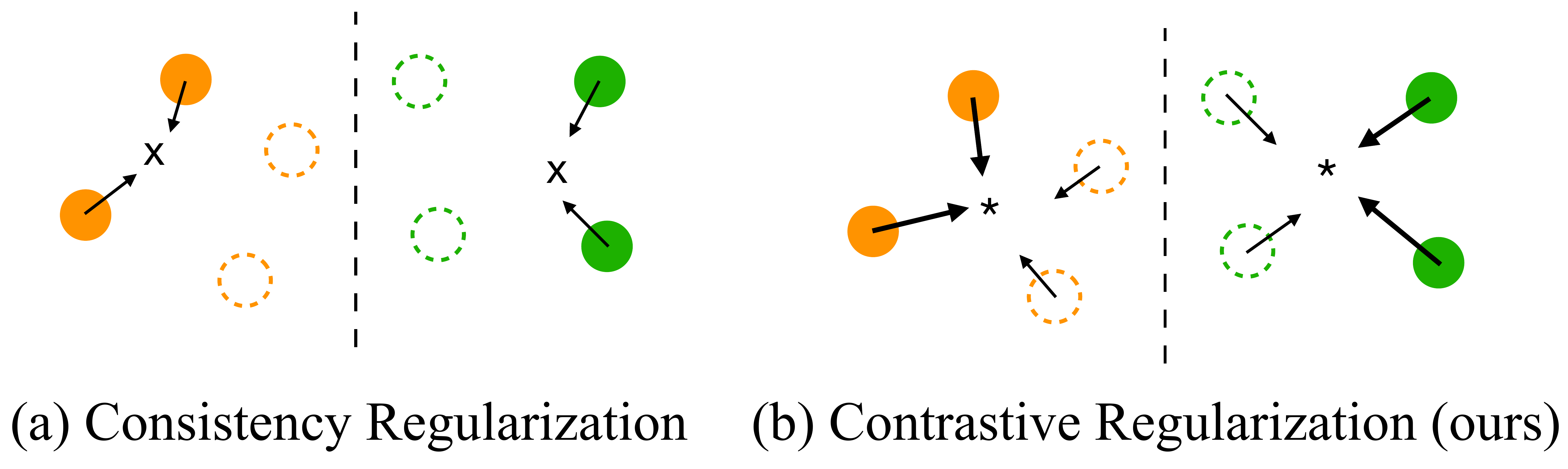}
\caption{The feature update of (a) consistency and (b) contrastive regularization. Different colors represent pseudo-labels. The circles with solid and dashed line are penultimate features having confident and unconfident pseudo-labels, respectively. The black dashed line is the decision boundary. The symbol $\times$ represents a cluster center that is estimated by confident samples only, and * represents a cluster center that is estimated by all samples in the same cluster. The length of arrows represents the magnitude of gradient vectors. The cluster centers are computed by the class weight vectors.}
\label{fig:concept}
\vspace{-0.1in}
\end{figure}

Based on the above analysis, we propose \textit{contrastive regularization} to improve the performance of SSL based on consistency regularization.
The main idea is described in Figure~\ref{fig:concept}.
The consistency regularization moves the features of strongly augmented samples having only confident pseudo-labels toward their corresponding class centers of the confident features by pseudo-labels. 
In contrast, the proposed contrastive regularization forms class clusters based on both confident and unconfident pseudo-labels.
Then, it moves the features having confident pseudo-labels toward the center positions of their clusters, while pulling the features of samples with both confident and unconfident pseudo-labels in the same cluster and pushing the features in different clusters.
Thus, a model can learn well-clustered features of unlabeled data, enabling the confident labeling information to be propagated into more unlabeled samples during training.

In the experiments, we show that our contrastive regularization improves the performance of consistency regularization methods on various SSL benchmarks, including SVHN, CIFAR-10, CIFAR-100, STL-10, and ImageNet with limited labels.
Especially, different from the previous methods, we show that our method leverages unlabeled samples in the early stage of training and requires much fewer iterations for the outperformance.
We also demonstrate that the contrastive regularization achieves the robust performance on the task of open-set SSL, which is more realistic in that unlabeled data contains out-of-distribution samples \cite{oliver2018realistic,yu2020multi}.
Finally, we conduct an extensive ablation study to show that the contrastive regularization is valid and not highly sensitive to the selection of hyper-parameters.

Our main contributions can be summarized as follows. 
1) It is the first study to analyze the limitation of the consistency regularization with respect to the efficiency of SSL training.
2) We propose a simple yet powerful solution, the contrastive regularization, which consistently improves the SSL performance on different SSL benchmarks with fewer training iterations than the previous consistency regularization.
3) Contrastive regularization shows the robustness on the more realistic benchmark that includes out-of-distribution samples in the unlabeled dataset.

\section{Related Work}
\label{sec:RW}

\paragraph{Consistency Regularization for SSL.} Recent SSL methods use the consistency regularization \cite{laine2016temporal} and focus on the policies of stochastic data augmentations such as adversarial perturbations \cite{miyato2018virtual} or mixup \cite{berthelot2019remixmatch,berthelot2019mixmatch,zhang2018mixup}.
As the most simplified yet powerful framework of SSL, UDA and FixMatch \cite{xie2020unsupervised,sohn2020fixmatch} show that the simple combination of strong data augmentation such as RandAugment \cite{cubuk2020randaugment} and pseudo-labeling \cite{lee2013pseudo} can obtain high performance.
Thus, we focus on improving the consistency regularization, because they have shown state-of-the-art results compared with other SSL approaches \cite{shi2018transductive,iscen2019label}.

\paragraph{Semi-Supervised Learning with Self-Supervision.} The SSL performance can be improved when self-supervised learning is used with an auxiliary task for representation learning, and our contrastive regularization can be viewed as an auxiliary task for SSL.
S4L \cite{zhai2019s4l} demonstrates that auxiliary tasks such as rotation or exemplar self-supervision can improve the SSL performance.
For time-series classification, forecasting of the next-step value is used as an auxiliary task \cite{jawed2020self}. 
CoMatch \cite{li2020comatch} unifies pseudo-labeling, self-supervised learning, and graph-based SSL, using the graph contrastive learning and the pseudo-label smoothing with a large size of memory bank \cite{he2020momentum}.
Although both CoMatch and our method use a contrastive loss to regularize the unlabeled features, our method can be easily in tandem with the consistency regularization with minimal change for the contrastive loss.

\paragraph{Pretraining and Finetuning.}
Finetuning after pretraining on an upstream task is a solution for learning a task with scarce labels, when large-scale labeled or unlabeled datasets are available for the upstream task.
For instance, a model, which is pretrained on a large-scale labeled dataset, can be well transferred to downstream tasks \cite{kolesnikov2019large}.
but a negative transfer occurs when the target task is unrelated to the upstream domain or task \cite{zamir2018taskonomy}.
When a large-scale unlabeled data is available, a framework using both \textit{task-agnostic} pretraining and task-specific finetuning can become a strong SSL approach \cite{vincent2010stacked,he2020momentum,chen2020big}.
However, utilizing unlabeled samples in a task-specific way can outperform a task-agnostic approach without a large number of unlabeled samples.
Thus, we emphasize that \textit{task-specific} SSL methods are important because it is hard to collect a large number of unlabeled samples in the real world.

\section{Contrastive Regularization for Semi-Supervised Learning}
\label{sec:method}
In this section, we introduce our \textit{contrastive regularization} to improve the SSL performance of the consistency regularization.
We first formulate SSL and the consistency regularization, which is the most common approach and shows remarkable results with deep neural networks (DNNs).

\subsection{Problem Formulation}
We assume that a labeled dataset $D_L$ and an unlabeled dataset $D_U$ are given to train a model parameterized by $\theta$.
A mini-batch $\mathcal{B}$ consists of $B$ labeled samples $\mathcal{X}=\{(x_b, y_b)|(x_b,y_b) \in D_L \}_{b=1}^{B}$ and $\mu B$ unlabeled samples $\mathcal{U}=\{u_b | u_b \in D_U\}_{b=1}^{\mu B}$, where $\mu$ is the ratio of unlabeled samples to the labeled samples in a mini-batch.
The total loss $\mathcal{L}$ is minimized at each training iteration
\begin{equation} \label{eq:ssl_form}
    \mathcal{L}(\mathcal{B}) = \mathcal{L}_{L}(\mathcal{X}) + \lambda_u \mathcal{L}_{U}(\mathcal{U}),
\end{equation}
where $\mathcal{L}_L$ is a supervised loss, $\mathcal{L}_U$ is an unsupervised loss, and $\lambda_u$ is an unlabeled loss ratio.
Cross entropy is used for a supervised loss, and the type of $\mathcal{L}_U$ determines how to leverage the unlabeled samples.
For example, entropy regularization \cite{grandvalet2005semi} and pseudo-labeling enforce the predictions on unlabeled samples to have a low entropy, so that the decision boundary is located in the low-density area \cite{sajjadi2016regularization}.

For an unlabeled sample $u \in \mathcal{U}$, the label prediction $\hat{p}(y|u)=\text{softmax}[W^\top h_\theta (u)]$ is given by the model with $\theta$ comprising $K$-class weight matrix $W=[w_1, w_2, ..., w_K] \in \mathbb{R}^{H \times K}$, where $h_\theta (u) \in \mathbb{R}^{H}$ denotes the penultimate features.
We define a stochastic function of strong augmentation as $\alpha$, and the set of strongly augmented samples for an unlabeled mini-batch as $\mathcal{A}_m(\mathcal{U})=\{u'_{i} | u \in \mathcal{U}, u'_{i} = \alpha(u), 1 \leq i \leq m \}$, where $m$ is the number of augmented view per an unlabeled sample in the mini-batch.
Then, the consistency regularization, $\mathcal{R}_{CS}$, is defined as 
\begin{equation} \footnotesize \label{eq:consistency_regularization}
    \mathcal{R}_{CS}(\mathcal{U}) = \frac{1}{|\mathcal{A}_m (\mathcal{U})|} \sum_{u' \in \mathcal{A}_m(\mathcal{U})}
    \mathds{1}[\max q_{u} > \delta] H(\hat{q}_{u}, \hat{p} (y|u')),
\end{equation}
where $u'$ is the augmented sample of $u \in \mathcal{U}$, $\delta$ is a confidence threshold, and $H$ is the cross entropy loss.
\textcolor{black}{In the remaining parts of this paper, a strongly augmented sample of $u$ is represented as $u'$ for brevity.}
$\hat{q}_{u}$ is the pseudo-label of $u'$ and defined as $\hat{q}_{u}= \argmax q_{u}$, where $q_{u}= {\text{sg}[\hat{p}(y|u)}]$ and sg is the stop gradient.
Note that the pseudo-label of $u'=\alpha(u)$ is determined by the label prediction on the sample without strong augmentation, $u$, for the reliability of pseudo-labeling.

The performance of consistency regularization highly depends on the choices of $\alpha$ and $\delta$.
Data augmentation encourages DNNs to learn the generalized representations with the local geometry of the data-manifold, assuming that the learned manifolds of different classes are well-separated \cite{verma2019interpolation,ghosh2021data}. 
Therefore, the features having different pseudo-labels become well-separated, propagating the confident (pseudo-)labeling information into their neighbors on the data manifold.
Determining the threshold $\delta$ is an inherent trade-off for the SSL performance, because the $\delta$ controls the balance of the reliability and the number of unlabeled samples leveraged.
A higher value of $\delta$ is commonly used to avoid a confirmation bias, but it restricts unlabeled samples to be included in SSL training and can preclude the model from learning the transformation-invariant representations on the excluded samples \cite{arazo2020pseudo}.
It minimizes the entropy of only a sample using a confident pseudo-labeling for SSL training.
\vspace{-0.15in}

\subsection{Training Inefficiency of Consistency Regularization} \label{sec:grad_cs}
As the consistency regularization achieves high SSL performance competitive with the fully supervised setting, it requires a large number of training iterations even on small-scale datasets.
For example, FixMatch \cite{sohn2020fixmatch} requires over 10,000 epochs to train WRN-28-2 \cite{zagoruyko2016wide} on the CIFAR-10 dataset.
However, when the labels are fully provided, about 100 epochs are enough to learn the dataset under supervision.

Here, we analyze the consistency regularization to show its training inefficiency.
Assume that $\hat{Q}_i$ is a set of strongly augmented samples assigned to the $i$-th class by the pseudo-label, $\hat{Q}_i = \{ u' | u' \in \mathcal{A}_m (\mathcal{U}), u \in \mathcal{U}, \hat{q}_{u} = i \} $.
The minus gradients of $\mathcal{R}_{CS}$ with respect to the features $h_\theta$ and to the $i$-th class weight vector $w_i$ are as follows:
\begin{equation} \footnotesize
    - \frac{\partial \mathcal{R}_{CS}}{\partial w_i} = \frac{1}{|\mathcal{A}_m (\mathcal{U})|}  \sum_{u' \in \hat{Q}_{i}} \mathds{1}[\max q_{u} > \delta] 
    h_{\theta}(u') (1-\hat{p}(i|u')), \label{eq:cs_class}
\end{equation}
\begin{equation} \footnotesize
\label{eq:cs_hidden}
\begin{aligned} 
    - \frac{\partial \mathcal{R}_{CS}}{\partial h_\theta} = \frac{1}{|\mathcal{A}_m (\mathcal{U})|} \sum_{u' \in \mathcal{A}_m(\mathcal{U})} & \mathds{1}[\max q_{u} > \delta]
    \{\sum_{i \neq {\hat{q}_{u}}} w_i \hat{p}(i|u')\\
    &+  w_{\hat{q}_{u}}(1-\hat{p}({\hat{q}_{u}}|u')) \}. 
\end{aligned}
\end{equation}
By this gradient analysis, we postulate that the inefficiency of the consistency regularization results from the exclusion of samples with unconfident pseudo-labels and the training bias on the confident pseudo-labels by the masking $\mathds{1}[\max q_{u} > \delta]$.
Figure~\ref{fig:concept}(a) contains the interpretation of the gradient analysis.
Here, we assume that the features with unconfident pseudo-labels are close to the decision boundary, considering the linearity of softmax classifier \cite{bishop2006pattern}.
The class weight vector $w_i$ is updated to the weighted sum of only confident features in Eq.~(\ref{eq:cs_class}).
Then, the confident features in Eq.~(\ref{eq:cs_hidden}) are updated by the class weight vectors, so the features only having confident pseudo-labels become close together.
However, the unconfident samples are excluded in the gradients computations, and the labeling information of confident samples cannot be effectively propagated into the unlabeled samples.
In addition, the class weight vector is slowly changed due to the exclusion of unconfident samples, because the gradient in Eq.~\ref{eq:cs_class} is bounded by the confidence threshold, $- \frac{\partial \mathcal{R}_{CS}}{\partial w_i} < \frac{1}{|\mathcal{A}_m (\mathcal{U})|} \sum_{u' \in \hat{Q}_{i}} \mathds{1}[\max q_{u_b} > \delta] h_{\theta}(u') (1-\delta)$, where $\delta$ is typically selected as a high value such as 0.95.
Thus, the model cannot leverage lots of unlabeled samples over the SSL training and requires a large number of training iterations to gradually increase the number of confident samples. 

\subsection{Contrastive Regularization for SSL} \label{sec:grad_cr}
We propose \textit{contrastive regularization} in Figure~\ref{fig:concept}(b) to effectively leverage unlabeled samples for SSL.
Even though Figure~\ref{fig:concept}(b) describes the two-class classification, the concept can be generalized to the setting of multiple classes, and all experiments in this study are also conducted on multi-class tasks.
Different from the consistency regularization, the class clusters are formed by the features with both confident and unconfident pseudo-labels.
Then, our method regularizes the hidden features of confident unlabeled samples to be moved toward the samples with unconfident pseudo-labels in the same cluster, and propagates the labeling information.
At the same time, to leverage the unconfident samples without decreasing the confident threshold $\delta$, the features having confident pseudo-labels pull the features of unconfident samples in the same cluster, while pushing the features in different clusters.
It can achieve the entropy minimization for SSL, and unlabeled samples are beneficial with a small overlap of classes, since the contrastive regularization can learn well-clustered features that reduce the overlaps.

For this, we modify SupContrast \cite{khosla2020supervised}, which is used for a supervised pretraining on large-scale labeled data, into SSL setting by adding a projection head after the penultimate features.
We define the set of \textit{pseudo}-positive pairs of $u'$ as $\hat{P}(u')=\{ p' | p' \in \mathcal{A}_m(\mathcal{U})/u', \hat{q}_p = \hat{q}_u \}$, where $\hat{q}_p$ and $\hat{q}_u$ are the pseudo-label of $p'$ and $u'$, respectively.
Note that a pseudo-label of a strongly augmented sample $u'$ is defined by the label prediction on the unlabeled sample $u$ before strong augmentation.
\textcolor{black}{The positive sample pairs represent the samples whose pseudo-labels are the same, and the augmented samples in $\hat{P}(u')$ have the same pseudo-label with $u'$.}
Then, the contrastive regularization, $\mathcal{R}_{CR}$, is defined as follows:
\begin{equation} \footnotesize
\mathcal{R}_{CR} (\mathcal{U}) = \frac{1}{|\mathcal{A}_m(\mathcal{U})|}\sum_{u'\in \mathcal{A}_m(\mathcal{U})} \mathds{1}[\max q_{u} > \delta'] r(u'), 
\label{eq:batch_CR}    
\end{equation}
\begin{equation} \footnotesize
r (u') = \frac{-1}{|\hat{P}(u')|} \sum_{p' \in \hat{P}(u')} \log \frac{\exp ({\langle z_{u'}, z_{p'} \rangle}{/\tau} )}{\sum_{v' \in \mathcal{A}_m(\mathcal{U})/u'} \exp ( {\langle z_{u'}, z_{v'} \rangle}{/\tau} )}, \label{eq:sample_CR}
\end{equation}
where $\delta'$ is a confidence threshold, $\tau$ is a temperature scaling parameter, and $z_{u'}$ is a \textit{normalized} vector of the projection head.
Our total loss is 
$\mathcal{L}(\mathcal{B}) = \mathcal{L}_L (\mathcal{X}) + \lambda_{CS} \mathcal{R}_{CS} (\mathcal{U}) + \lambda_{CR} \mathcal{R}_{CR} (\mathcal{U})$, where $\lambda_{CS}$ and $\lambda_{CR}$ is the loss ratio of consistency and contrastive regularization, respectively.

The features of confident samples move toward the centroid of its feature cluster, which consists of features having the same pseudo-labels, and pull the unconfident features in the same cluster by our contrastive regularization.
Without the loss of generalizability, we notate the softmax score of $z_{p'}$ with $z_{u'}$ as $s[u',p']$, and assume the normalized vector $z=h$, and $\tau=1$.
For $u'$ and $v' \in \mathcal{A}_m (\mathcal{U})/u'$, the minus gradients of $r(u')$ with respect to $h_\theta$ are as follows: 
\begin{equation} \footnotesize
    - \frac{\partial r(u')}{\partial h_\theta(u')} = \sum_{p'\in \hat{P}(u')} (\frac{1}{|\hat{P}(u')|} - s[u', p']) h_\theta(p') + R(u'), \label{eq:gradient_cr}
\end{equation}
\begin{equation} \label{eq:gradient_cr_unconf}  \footnotesize
    -\frac{\partial r(u')}{\partial h_\theta(v')} = 
    \begin{cases}
     (\frac{1}{|\hat{P}(u')|} - s[u',v'])h_\theta(u'), &  \mbox{if } v' \in \hat{P}(u') \\
     ~~~~ - s[u',v']h_\theta(u'), & \mbox{if } v' \notin \hat{P}(u')
    \end{cases},
\end{equation}
where $R(u')$ is a remainder term and small enough.
We attach the detailed derivation of Eq.~(\ref{eq:gradient_cr}) and~(\ref{eq:gradient_cr_unconf}) in Appendix A.
If the $u'$ has a confident pseudo-label as Eq.~(\ref{eq:gradient_cr}), the contrastive regularization updates its feature vector $h_\theta (u')$ toward the weighted sum of positive features both with confident and unconfident pseudo-labels.
Different from the consistency regularization, the feature update of confident samples also considers the features with unconfident pseudo-labels in the same cluster.
At the same time, in Eq.~(\ref{eq:gradient_cr_unconf}), the confident features pull the features of both confident and unconfident samples in the same cluster $\hat{P}(u')$, while pushing the features in different clusters.
Although our contrastive regularization of a confident feature $u'$ learns to aggregate positive samples with $s[u',v']=1/|\hat{P}(u')|$, the $u'$ can push a positive sample $v'$ of $u'$ with a negative value in Eq.~(\ref{eq:gradient_cr_unconf}) \textit{during training}, because all positive samples in a mini-batch are included in the denominator of the long term in Eq.~(\ref{eq:sample_CR}).
However, note that other negative samples in the different clusters still push $v'$, avoiding a wrong cluster assignment by the negative values of Eq.~(\ref{eq:gradient_cr_unconf}) during training.
Thus, the model can propagate the confident labeling information into the unlabeled samples, while learning well-clustered features for SSL \cite{castelli1996relative}.

Although our contrastive regularization utilizes the information of unconfident pseudo-labeling, the confirmation bias does not more increase than previous consistency regularization methods.
According to Appendix C, the performance degradation by the memorization of wrong pseudo-labels occurs in the later stage of SSL training.
In the early stage of training, our method learns well-clustered representations of unlabeled samples to effectively propagate labeling information of labeled samples and unlabeled samples with confident pseudo-labeling.
Thus, the contrastive regularization can improve the SSL performance before the SSL model starts to memorize wrong pseudo-labels~\cite{arpit2017closer}.
In addition, different from the consistency regularization, our method is performed on features of unlabeled samples, not directly on class predictions, to avoid the memorization of wrong labels by the contrastive regularization.

\begin{table*}[]
\centering
\footnotesize
\caption{Test accuracies (\%) for SVHN and CIFAR-10 on five different runs with randomly selected labeled samples. The Asterisks mean that the results are from the previous studies \cite{sohn2020fixmatch,li2020comatch,kim2021selfmatch}.}
\label{tab:cifar_acc}
\begin{tabular}{l|cccc|cccc}
\toprule
 & \multicolumn{4}{c|}{SVHN} & \multicolumn{4}{c}{CIFAR-10} \\ \hline
 Method & 20 labels & 40 labels & 250 labels & 1000 labels&  20 labels & 40 labels & 250 labels & 4000 labels       \\ \hline
 MixMatch*           &   - & 57.45\scriptsize{$\pm$14.53} & 96.02\scriptsize{$\pm$0.23} & 96.50\scriptsize{$\pm$0.28} & -  &  52.46\scriptsize{$\pm$11.50} & 88.95\scriptsize{$\pm$0.86} & 93.58\scriptsize{$\pm$0.10}  \\
 UDA*           &   - & 43.75\scriptsize{$\pm$20.51} & 94.31\scriptsize{$\pm$2.76} & 97.54\scriptsize{$\pm$0.24} & -  &  70.95\scriptsize{$\pm$5.93} & 91.18\scriptsize{$\pm$1.08}  & 95.12\scriptsize{$\pm$0.18} \\
 ReMixMatch*    &   -  & \textbf{96.66\scriptsize{$\pm$0.20}} & 97.08\scriptsize{$\pm$0.48} & 97.35\scriptsize{$\pm$0.08} & - & 81.90\scriptsize{$\pm$9.64} & 94.46\scriptsize{$\pm$0.05} & 95.28\scriptsize{$\pm$0.13}  \\ 
 CoMatch*    &   -  & - & - & - & 81.85\scriptsize{$\pm$5.56} & 91.51\scriptsize{$\pm$2.15} & -  & -  \\ \hline
 FixMatch    & 90.05\scriptsize{$\pm$8.01} & 94.83\scriptsize{$\pm$2.24} & 97.28\scriptsize{$\pm$0.66} & 97.46\scriptsize{$\pm$0.09} & 74.98\scriptsize{$\pm$11.38}  & 91.24\scriptsize{$\pm$3.72} & 94.67\scriptsize{$\pm$0.28} &  95.57\scriptsize{$\pm$0.05}\\
 FixMatch+CR   & \textbf{94.96\scriptsize{$\pm$4.77}} & 96.33\scriptsize{$\pm$1.84} & \textbf{97.55\scriptsize{$\pm$0.08}} & \textbf{97.61}\scriptsize{$\pm$0.06} & \textbf{88.26}\scriptsize{$\pm$1.38} & \textbf{94.31}\scriptsize{$\pm$0.90} & \textbf{94.96}\scriptsize{$\pm$0.30} & \textbf{95.84}\scriptsize{$\pm$0.13}\\ \hline
 SelfMatch*    &   -  & 96.58\scriptsize{$\pm$1.02} & 97.37\scriptsize{$\pm$0.43} & 97.49\scriptsize{$\pm$0.07} & - & 93.19\scriptsize{$\pm$1.08} & 95.13\scriptsize{$\pm$0.26} & 95.94\scriptsize{$\pm$0.08}  \\ 
 FixMatch+CR++   & \textbf{96.88}\scriptsize{$\pm$0.60} & \textbf{97.05}\scriptsize{$\pm$0.28} & \textbf{97.95}\scriptsize{$\pm$0.09} & \textbf{98.11}\scriptsize{$\pm$0.05}& \textbf{94.24}\scriptsize{$\pm$3.48} & \textbf{95.26}\scriptsize{$\pm$0.70} & \textbf{96.00}\scriptsize{$\pm$0.31} & \textbf{96.68}\scriptsize{$\pm$0.18}\\
\bottomrule
\end{tabular}
\end{table*}

\section{Experiments} \label{sec:EXP}
We empirically validate that the contrastive regularization consistently improves the performance on standard SSL benchmarks such as SVHN, CIFAR-10, CIFAR-100, STL-10, and ImageNet with limited labels.
We also show that the contrastive regularization is also robust to the open-set SSL setting, and an extensive ablation study is conducted in this section.
For experiments, we use an exponential moving average (EMA) of model parameters \cite{tarvainen2017mean} with 0.999 momentum and cosine learning rate scheduling in \cite{sohn2020fixmatch} for all experiments.
The training epochs are computed based on the batch size of unlabeled samples.
For a fair comparison, we follow the experimental setting in the previous study \cite{sohn2020fixmatch}, and attach the implementation details in Appendix B.

\subsection{Classification of SVHN, CIFAR-10, CIFAR-100}
To analyze the effect of contrastive regularization (CR), we reproduce FixMatch and UDA using Pytorch 1.6.0 \cite{NEURIPS2019_9015}.
For a fair comparison with previous studies, we use the encoder of WRN-28-2 (1.5M parameters) for SVHN and CIFAR-10, and a WRN-28-8 (23.4M parameters) for CIFAR-100.
For SVHN and CIFAR-10, we also use WRN-28-8 (FixMatch+CR++) for comparison with SelfMatch \cite{kim2021selfmatch} which uses over 21M parameters.

For the projection embedding $z$, we add a 2-layer MLP after the feature extractor $h_\theta$, and its dimension sizes are 64 for WRN-28-2 and 256 for WRN-28-8.
We use RandAugment \cite{cubuk2020randaugment} for strong data augmentation, and set $\lambda_{CS}=1.0$.
We use $\lambda_{CR}=1.0$ for SVHN and CIFAR-10, and $\lambda_{CR}=10.0$ for CIFAR-100.
Following \cite{sohn2020fixmatch}, FixMatch and UDA use 10,500 epochs of unlabeled data.
FixMatch+CR uses 6,500 epochs for CIFAR-10 and SVHN, and 2,500 epochs for CIFAR-100 to achieve state-of-the-art results.
Nevertheless, note that much less time needs to outperform FixMatch in the next section.

For SVHN and CIFAR-10, Table~\ref{tab:cifar_acc} shows that our method consistently improves the SSL performance of FixMatch on the same codebase.
Consequently, the proposed FixMatch+CR achieves the state-of-the-art performance of WRN-28-2 except SVHN with 40 labels.
Although FixMatch+CR cannot outperform the reported result of ReMixMatch \cite{berthelot2019remixmatch} on SVHN with 40 labels, our method improves the test accuracy of FixMatch by 1.50\%.

\begin{table} 
\footnotesize
\centering
\caption{Test accuracy (\%) of WRN-28-8 on the CIFAR-100 dataset with 400, 2500, and 10000 labels.}
\begin{tabular}{l|ccc}
\hline
 & \multicolumn{3}{c}{CIFAR-100} \\ \hline
Medthod     & 400 labels & 2500 labels & 10000 labels \\ \hline
UDA & 48.02\scriptsize{$\pm$2.66} & 70.50\scriptsize{$\pm$0.53} & 77.07\scriptsize{$\pm$0.33} \\
UDA+CR & \textbf{49.91}\scriptsize{$\pm$0.79} & \textbf{72.12}\scriptsize{$\pm$0.28} & \textbf{78.58}\scriptsize{$\pm$0.11} \\ \hline
FixMatch & 48.48\scriptsize{$\pm$0.55} & 71.53\scriptsize{$\pm$0.29} & 78.03\scriptsize{$\pm$0.26} \\
FixMatch+CR & \textbf{50.77}\scriptsize{$\pm$0.79} & \textbf{72.42}\scriptsize{$\pm$0.37} & \textbf{78.97}\scriptsize{$\pm$0.23} \\ \hline
\end{tabular}
\label{tab:ssl_cifar100}
\vspace{-0.1in}
\end{table}

We emphasize that the contrastive regularization has remarkable performance gains.
For the setting of 20 labels (2 labels per class), FixMatch+CR significantly improves the accuracy and the robustness to the selection of labeled samples.
FixMatch+CR outperforms CoMatch \cite{li2020comatch}, which has firstly reported the results on CIFAR-10 with 20 and 40 labels.
In addition, WRN-28-2 with FixMatch+CR are competitive with SelfMatch, although the number of parameters is about 15 times smaller.
When we increase the number of parameters into 23.4M, FixMatch+CR++ outperforms SelfMatch in all experimental settings of SVHN and CIFAR-10.

Our method is also effective on the CIFAR-100 dataset with 400, 2,500, and 10,000 labels (Table~\ref{tab:ssl_cifar100}).
When the contrastive regularization is used along with UDA and FixMatch, it improves the performance and outperforms the previous methods.
Note that the performance gains are significant and consistent regardless of the number of labels.

\subsection{Classification of STL-10 and ImageNet}

We evaluate our contrastive regularization on a larger scale of datasets such as STL-10 and ImageNet.
We set $\lambda_{CS}=1.0$ for SVHN and $\lambda_{CS}=10.0$ for ImageNet, and $\lambda_{CR}=10$ for the two.
The STL-10 dataset includes 5,000 labeled and 100,000 unlabeled 96$\times$96 images in 10 classes.
We train WRN-37-2 (5.9M parameters) on STL-10 with five folds of 1,000 and 5,000 labels.
10,500 and 5,000 epochs are used for FixMatch and FixMatch+CR, respectively.
The projection head uses 2-layer MLP with 256 dimensions.
For 1,000 labels, FixMatch+CR improves the results of FixMatch in Table~\ref{tab:ssl_stl_imagenet}.
For 5,000 labels, FixMatch+CR achieves 95.40\%, improving 95.18\% of FixMatch.

\begin{table}
\footnotesize
\centering
\caption{Test accuracy (\%) on the STL-10 and ImageNet datasets. Top-1 (top-5) accuracies are reported for ImageNet}
\begin{tabular}{l|c|cc}
\hline
 & STL-10 & \multicolumn{2}{c}{ImageNet} \\ \hline
Method     & 1,000 labels & 1\% labels & 10\% labels \\ \hline
FixMatch & 89.34\scriptsize{$\pm$1.79} & 51.29 (72.48) & 72.18 (89.98) \\
FixMatch+CR & \textbf{93.04}\scriptsize{$\pm$0.42} & \textbf{57.77 (78.12)} & \textbf{72.77 (90.15)} \\ \hline
\end{tabular}
\label{tab:ssl_stl_imagenet}
\vspace{-0.15in}
\end{table}

We also evaluate our method on the ImageNet dataset that includes about 1.3M training images in 1,000 object classes.
We use a self-supervised and pretrained ResNet-50 model by MoCo v2 \cite{he2020momentum,chen2020improved}, since reproducing FixMatch on ImageNet from scratch requires expensive cost such as about three days using 32 cores of TPU.
We use 1,024 labeled and 5,120 unlabeled images in each mini-batch, and train a model in 300 epochs of unlabeled data.
2-layer MLP with 512 dimensions is used for the projection embedding.
For 1\% and 10\% of labels, our contrastive regularization improves the accuracy of FixMatch in Table~\ref{tab:ssl_stl_imagenet}, and our method significantly improves the performance on the fewer labels.
Thus, we conclude that our contrastive regularization is also effective on large-scale datasets.

\begin{figure}
\centering
\includegraphics[width=3.315in]{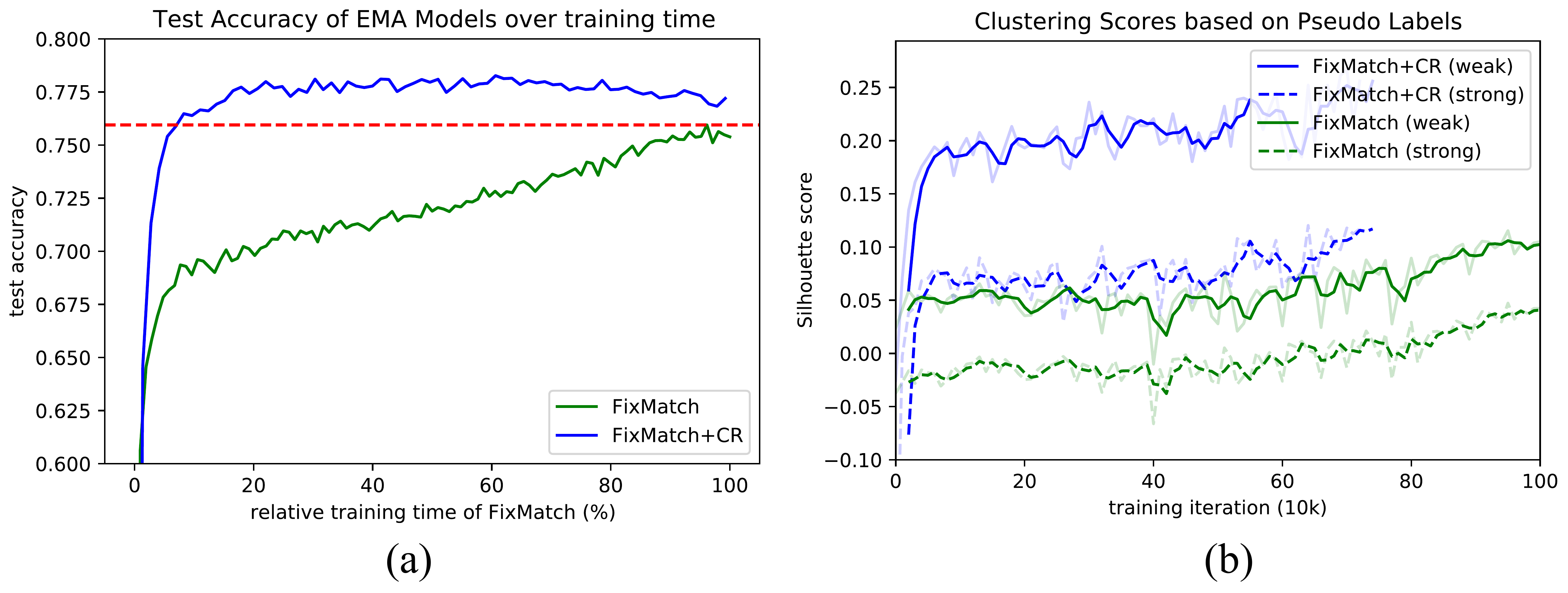}
\caption{Results of FixMatch and FixMatch+CR with WRN-28-4 trained on the CIFAR-100 with 10000 labels. (a) Test accuracy over training time, (b) Silhouette score of penultimate features based on pseudo-labels.}
\label{fig:emp_c100_w4}
\vspace{-0.1in}
\end{figure}

\subsection{Cost Efficiency of the Contrastive Regularization} \label{sec:efficiency}

The contrastive regularization not only improves the accuracy, but also enhances the training efficiency of SSL.
For a fair comparison of training time, four NVIDIA V100 GPUs are used to train both FixMatch and FixMatch+CR.
In Figure~\ref{fig:emp_c100_w4}(a), the accuracy of FixMatch gradually increases over the entire training time of 10,500 epochs.
Although one iteration of FixMatch+CR takes about 1.5$\times$ more time than FixMatch due to the use of two strongly augmented views, FixMatch+CR only takes 31\% of the total training time of FixMatch to achieve the best performance.
Also, 7\% of the training time of FixMatch is enough for FixMatch+CR to achieve the best performance of FixMatch (dashed line).
For other datasets, 2,500 epochs for SVHN and CIFAR-10, 1,000 epochs for CIFAR-100, and 1,500 epochs for STL-10 are enough to outperform FixMatch of 10,500 epochs, as shown in Appendix C.
Consequently, our method can save the training cost, reducing the training time and iterations.

We conjecture that the improved efficiency comes from the well-clustered representations by the contrastive regularization in the early stage of training.
Figure~\ref{fig:emp_c100_w4}(b) shows how features are well-clustered according to their pseudo-labels in terms of Silhouette score \cite{rousseeuw1987silhouettes}. 
If the decision boundary lies in the low-density regions and the features are well-clustered, the score is closed to +1, otherwise it is closed to -1.
For the features of strongly augmented samples, the clustering scores of FixMatch are near zero and it increases slowly after 40K iterations.
However, the clustering score of FixMatch+CR increases fast in the early stage of training. 
In addition, the scores are much higher than those of FixMatch during the entire training. 
This means that our contrastive regularization is effective in feature clustering, especially in the early training stage, and eventually improves both the training efficiency and final performance. 

\begin{table} 
\centering
\footnotesize
\caption{Test accuracy (\%) with different sizes of the widen factor on the same random seed. 28 layers of WRN is trained on CIFAR-100 with 2500 labels.}
\label{tab:abl_wf}
\begin{tabular}{l|cccc}
\toprule
 Widen Factor & 1 & 2 & 4 & 8 \\ \hline
 \# of Params & 0.38M & 1.48M & 5.87M & 23.40M \\ \hline
FixMatch & 55.86 & 64.74 & 69.75 & 72.02 \\
FixMatch+CR & \textbf{59.94} & \textbf{69.03} & \textbf{72.00} & \textbf{72.83} \\
\bottomrule
\end{tabular}
\vspace{-0.1in}
\end{table}

Table~\ref{tab:abl_wf} shows that the contrastive regularization is especially effective to a smaller model for SSL.
When the contrastive regularization is applied to WRN-28-4 and WRN-28-2 with 2500 labels of the CIFAR-100, the accuracies are improved by 2.25\% and 4.29\%, respectively.
Thus, the obtained accuracies of WRN-28-2 and WRN-28-4 with the contrastive regularization are comparable with those of WRN-28-4 and WRN-28-8 without it, respectively.
Note that increasing the widen factor by two times leads to a four times larger number of trainable parameters.

\subsection{Open-Set Semi-Supervised Learning}
For a realistic evaluation, open-set SSL \cite{oliver2018realistic,yu2020multi} assumes that an unlabeled dataset includes out-of-distribution (OOD) samples, which are totally different from the training and test samples.
Considering SVHN and CIFAR-10 as OOD of CIFAR-100, we add the OOD samples to the unlabeled data of CIFAR-100, and train WRN-28-4 on CIFAR-100 with 2500 and 10000 labels. 
Then, we evaluate the accuracy on the test data of CIFAR-100 according to the number of added OOD samples such as 10K, 20K, 30K, and 40K.

\begin{figure}
\centering
\includegraphics[width=3.315in]{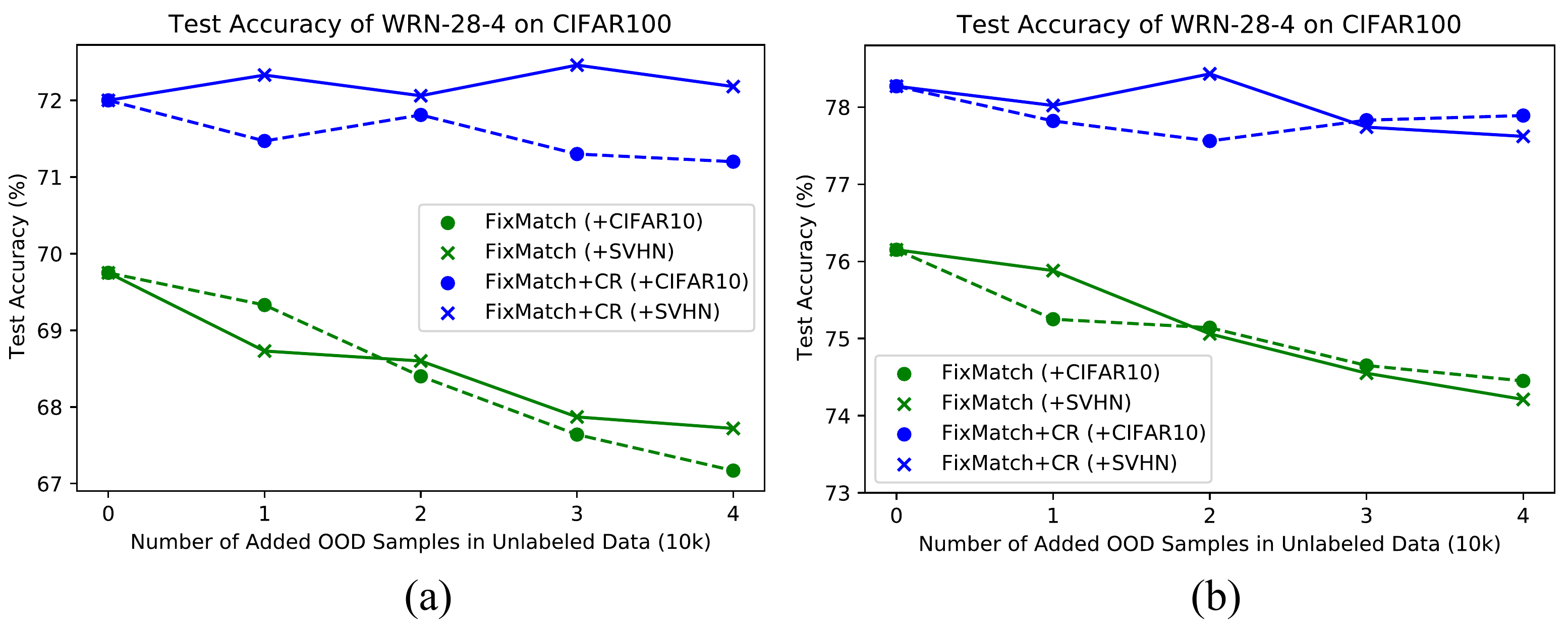}
\caption{Open-set SSL results of WRN-28-4 on CIFAR-100 with (a) 2500 and (b) 10000 labels. OOD samples (SVHN, CIFAR-10) are added into the unlabeld samples.}
\label{fig:opensetSSL}
\vspace{-0.125in}
\end{figure}

As shown in Figure~\ref{fig:opensetSSL}, the contrastive regularization enhances the robustness of SSL to the OOD samples.
FixMatch has severe degradation of accuracy as OOD samples are added into unlabeled data.
However, FixMatch+CR avoids the accuracy degradation and always outperforms FixMatch regardless of the number of OOD samples, because FixMatch+CR effectively leverages unlabeled samples from in-distribution, when the number of labels is limited. (Table~\ref{tab:cifar_acc}).
Note that FixMatch+CR is more robust to the OOD samples from SVHN than those from CIFAR-10, since SVHN has a totally different class distribution from CIFAR-100 and less affects the discrimination of the CIFAR-100 classes.

\subsection{Ablation Study}

\begin{figure*}
\centering
\includegraphics[width=6.5in]{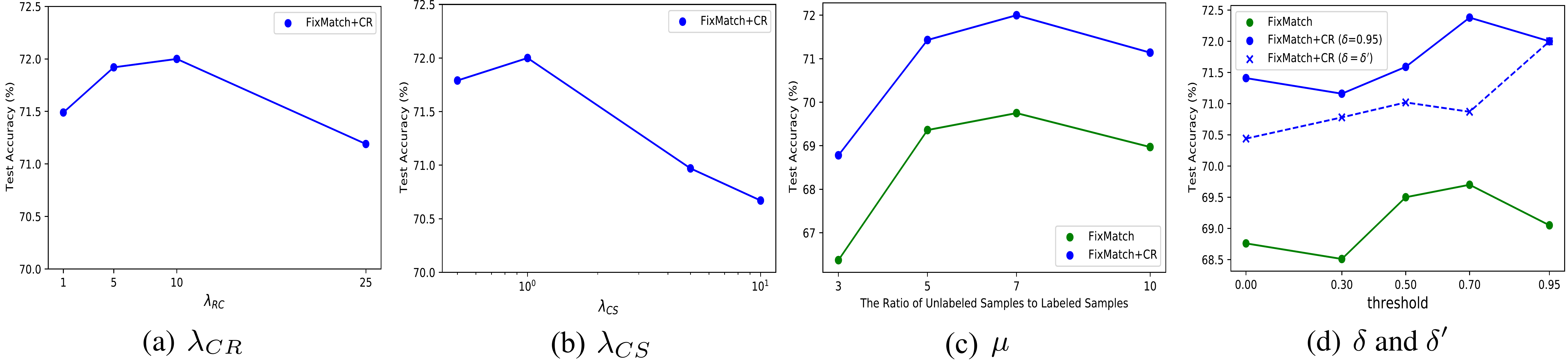}
\caption{The ablation study on the hyper-parameters (WRN-28-4, CIFAR-100 with 2500 labels): (a) contrastive regularization loss ratio, \textcolor{black}{(b) consistency regularization loss ratio, (c) unlabeled sample ratio in a mini-batch,} and (d) confidence thresholds.}
\label{fig:hyper_abl}
\end{figure*}

\textbf{Effects of Hyper-parameter Settings.}
We conduct an extensive ablation study to understand the effects of the different components in our method.
In Figure~\ref{fig:hyper_abl}(a), the contrastive regularization improves the accuracy when the weight of the contrastive loss $\lambda_{CR}$ is large enough ($\lambda_{CS}$ is fixed to $1.0$). 
Although an excessive large value of $\lambda_{CR}$ deteriorates the test accuracy by increased confirmation bias, the performance is robust to the selection of the $\lambda_{CR}$.

\textcolor{black}{
Figure~\ref{fig:hyper_abl}(b) shows the effect of the consistency regularization on the SSL performance, where the weight of contrastive loss $\lambda_{CR}$ is fixed to $10.0$. 
The test accuracy decreases when the weight of the consistency regularization $\lambda_{CS}$ increases, since the relative effect of our contrastive regularization decreases.
When the $\lambda_{CS}$ becomes smaller than 1.0, the test accuracy is competitive with $\lambda_{CS}=1.0$ and shows the effectiveness of our method.
}

\textcolor{black}{
Although the consistency regularization can completely be replaced with our contrastive regularization, we use the two regularizations to consistently achieve the state-of-the-art performance on different settings of the number of labeled samples.
In Table~\ref{tab:cs_cr}, we remove the consistency regularization and evaluate our contrastive regularization alone (CR-only, $\lambda_{CS}=0$) on CIFAR-10 with WRN-28-2.
When 4000 labels are available, CR-only still outperforms FixMatch, but the performance of CR-only is degraded when 40 and 250 labels are used.
Without the consistency regularization, the task-specific classification head is trained only by a supervised loss on labeled data $\mathcal{L}_L(\mathcal{X})$ in Eq.~(\ref{eq:ssl_form}), and it can be easily overfitted when the number of labeled samples is few.
Thus, our method is complementary to the consistency regularization to maximize the SSL performance.
}

In Figure~\ref{fig:hyper_abl}(c), the accuracies of both FixMatch and FixMatch+CR decrease when the ratio of unlabeled samples is low.
This observation is consistent with the findings in UDA and FixMatch.
It indicates that both consistency and contrastive approaches require a sufficiently large number of unlabeled samples in a mini-batch for high SSL performance.


Figure~\ref{fig:hyper_abl}(d) shows that the confident threshold is related to the trade-off between the reliability of pseudo-labeling and the number of unlabeled samples leveraged.
The confidence thresholds of the consistency and contrastive regularizations are denoted as $\delta$ in Eq.~(\ref{eq:consistency_regularization}) and $\delta'$ in Eq.~(\ref{eq:batch_CR}), respectively.
The low value of $\delta$ worsens the performance of both FixMatch and FixMatch+CR because of the low reliability of pseudo-labeling.
Although the test accuracy of FixMatch with $\delta=0$ drops to 68.74\%, only half of the training epochs are needed to achieve the performance, since it leverages all unlabeled samples regardless of the pseudo-labeling quality.
When $\delta'$ becomes low, FixMatch+CR suffers from the confirmation bias to some degree, but the test accuracy of FixMatch+CR with $\delta=\delta'=0$ still outperforms FixMatch, since our contrastive regularization effectively leverages unlabeled samples to improve the SSL performance.
When $\delta$ keeps high value (0.95) and $\delta'$ becomes low, the performance of FixMatch+CR decreases, but the results with $\delta'=0$ still significantly outperforms FixMatch with $\delta=0.95$.
The results imply that our contrastive regularization at the feature-level does not explicitly update the weights of the classifier, and therefore it shows robust results to the noisy pseudo-labels. 
However, as generating reliable pseudo-labels is still important to improve the SSL performance, we apply the selection mask with $\delta'>0$ in our method.
\textcolor{black}{Note that, different from the consistency regularization, our contrastive regularization in Eq.~(\ref{eq:gradient_cr_unconf}) can still leverage unlabeled samples with both confident and unconfident pseudo-labels, while keeping high reliability of pseudo-labeling by the confidence threshold.}

\begin{table} 
\centering
\footnotesize
\caption{Test accuracies (\%) on the numbers of views ($m$) per sample in $\mathcal{A}_m$ (CIFAR-100 with 10000 labels).}
\label{tab:abl_view}
\begin{tabular}{l|ccc}
\toprule
\# of Views  & $m=1$ & $m=2$ & $m=3$ \\ \hline
FixMatch & 76.01 & 76.03 & 76.67 \\
FixMatch+CR & 76.08 & \textbf{78.27} & 77.83 \\
\bottomrule
\end{tabular}
\end{table}

\begin{table} 
\centering
\caption{Test accuracies (\%) of contrastive regularization without consistency regularization on CIFAR-10}
\label{tab:cs_cr}
\footnotesize
\begin{tabular}{l|ccc}
\toprule
CIFAR-10  & 40 labels & 250 labels & 4000 labels \\ \hline
FixMatch & 94.81 & 95.11 & 95.6 \\
CR  & 91.51 & 94.41 & 95.89 \\
FixMatch+CR & \textbf{95.32} & \textbf{95.39} & \textbf{95.92} \\
\bottomrule
\end{tabular}
\vspace{-0.1in}
\end{table}


\textbf{Effect of the Number of Views.} \label{sec:num_views} 
FixMatch and FixMatch+CR are compared with different numbers of augmented views $m$ of $\mathcal{A}_m$.
Note that at least two views of each sample are required for FixMatch+CR to assure the existence of a positive sample in each mini-batch \cite{chen2020simple}.
Table~\ref{tab:abl_view} shows that the performance gain does not come from the solely increased number of views in our contrastive regularization.
The accuracy of FixMatch is not improved by two augmented views and does not outperform FixMatch+CR although three augmented views are used.
The results imply that the performance gain by our method does not depend on the increased number of views, but is from effectively leveraging more unlabeled data in SSL training shown in Figure~\ref{fig:emp_c100_w4}.

\textbf{Comparison with Unsupervised Contrastive Loss.}
We compare our method, which uses a contrastive loss with pseudo-labels for regularization, with the unsupervised contrastive loss (NT-Xent \cite{chen2020simple}).
As an auxiliary task, a self-supervised pretext task is known to improve the performance of semi-supervised learning.
Thus, we assume that unsupervised NT-Xent also improves the performance of semi-supervised learning, if the improvement by the contrastive regularization depends only on an auxiliary representation learning.
Table~\ref{tab:nt-xent} shows that the NT-Xent highly decreases the SSL performance. 
This might be due to that the task-agnostic instance discrimination tends to push away semantically similar instances in the same class.

\begin{table} 
\centering
\footnotesize
\caption{SSL performance of unsupervised contrastive learning as the SSL regularization.}
\label{tab:nt-xent}
\begin{tabular}{l|ccc}
\toprule
  & FixMatch & +NT-Xent & +CR \\ \hline
400 labels & 49.42 & 32.27 & \textbf{50.15} \\
2500 labels & 69.75&  54.44 & \textbf{72.00} \\
10000 labels & 76.15&  67.02 & \textbf{78.27} \\
\bottomrule
\end{tabular}
\vspace{-0.05in}
\end{table}

\section{Conclusion} \label{sec:conclusion}

Semi-supervised learning is important to learn a task with limited labels, while effectively leveraging unlabeled data in a task-specific way.
In this work, we show that the SSL training of the previous consistency regularization is biased on unlabeled samples only with confident pseudo-labeling by a selection mask.
Thus, we propose contrastive regularization that significantly improves SSL performance and can be used along with the consistency regularization by minimal change of implementation.
On various SSL benchmarks, the contrastive regularization not only improves the test accuracy but also significantly reduces the number of training iterations to achieve high performance.
Especially, our method shows more effective results on a dataset containing fewer labels or out-of-distribution samples.

In future work, 
\textcolor{black}{our approach can be applied to other SSL methods based on pseudo-labels~\cite{xie2020self,chen2020big} on large-scale datasets. However, our method still has a limitation on the memory efficiency for large-scale datasets due to large batch size, although the contrastive regularization improves training efficiency and the SSL performance.}
Thus, it is also worth exploration for leveraging large-scale unlabeled datasets in a task-specific way.
{

\section{Acknowledgements}
This work was supported by Institute of Information \& communications Technology Planning \& Evaluation(IITP) grant funded by the Korea government(MSIT) (No.2018-0-01398, Development of a Conversational, Self-tuning DBMS).
\bibliography{aaai22}

\begin{thebibliography}{37}
\providecommand{\natexlab}[1]{#1}

\bibitem[{Arazo et~al.(2020)Arazo, Ortego, Albert, O’Connor, and
  McGuinness}]{arazo2020pseudo}
Arazo, E.; Ortego, D.; Albert, P.; O’Connor, N.~E.; and McGuinness, K. 2020.
\newblock Pseudo-labeling and confirmation bias in deep semi-supervised
  learning.
\newblock In \emph{2020 International Joint Conference on Neural Networks
  (IJCNN)}, 1--8. IEEE.

\bibitem[{Arpit et~al.(2017)Arpit, Jastrzebski, Ballas, Krueger, Bengio,
  Kanwal, Maharaj, Fischer, Courville, Bengio et~al.}]{arpit2017closer}
Arpit, D.; Jastrzebski, S.; Ballas, N.; Krueger, D.; Bengio, E.; Kanwal, M.~S.;
  Maharaj, T.; Fischer, A.; Courville, A.; Bengio, Y.; et~al. 2017.
\newblock A closer look at memorization in deep networks.
\newblock In \emph{International Conference on Machine Learning}, 233--242.
  PMLR.

\bibitem[{Berthelot et~al.(2019{\natexlab{a}})Berthelot, Carlini, Cubuk,
  Kurakin, Sohn, Zhang, and Raffel}]{berthelot2019remixmatch}
Berthelot, D.; Carlini, N.; Cubuk, E.~D.; Kurakin, A.; Sohn, K.; Zhang, H.; and
  Raffel, C. 2019{\natexlab{a}}.
\newblock ReMixMatch: Semi-Supervised Learning with Distribution Matching and
  Augmentation Anchoring.
\newblock In \emph{International Conference on Learning Representations}.

\bibitem[{Berthelot et~al.(2019{\natexlab{b}})Berthelot, Carlini, Goodfellow,
  Papernot, Oliver, and Raffel}]{berthelot2019mixmatch}
Berthelot, D.; Carlini, N.; Goodfellow, I.; Papernot, N.; Oliver, A.; and
  Raffel, C.~A. 2019{\natexlab{b}}.
\newblock MixMatch: A Holistic Approach to Semi-Supervised Learning.
\newblock \emph{Advances in Neural Information Processing Systems}, 32.

\bibitem[{Bishop(2006)}]{bishop2006pattern}
Bishop, C.~M. 2006.
\newblock \emph{Pattern recognition and machine learning}.
\newblock springer.

\bibitem[{Castelli and Cover(1996)}]{castelli1996relative}
Castelli, V.; and Cover, T.~M. 1996.
\newblock The relative value of labeled and unlabeled samples in pattern
  recognition with an unknown mixing parameter.
\newblock \emph{IEEE Transactions on information theory}, 42(6): 2102--2117.

\bibitem[{Chen et~al.(2020{\natexlab{a}})Chen, Kornblith, Norouzi, and
  Hinton}]{chen2020simple}
Chen, T.; Kornblith, S.; Norouzi, M.; and Hinton, G. 2020{\natexlab{a}}.
\newblock A simple framework for contrastive learning of visual
  representations.
\newblock In \emph{International conference on machine learning}, 1597--1607.
  PMLR.

\bibitem[{Chen et~al.(2020{\natexlab{b}})Chen, Kornblith, Swersky, Norouzi, and
  Hinton}]{chen2020big}
Chen, T.; Kornblith, S.; Swersky, K.; Norouzi, M.; and Hinton, G.~E.
  2020{\natexlab{b}}.
\newblock Big Self-Supervised Models are Strong Semi-Supervised Learners.
\newblock \emph{Advances in Neural Information Processing Systems}, 33:
  22243--22255.

\bibitem[{Chen et~al.(2020{\natexlab{c}})Chen, Fan, Girshick, and
  He}]{chen2020improved}
Chen, X.; Fan, H.; Girshick, R.; and He, K. 2020{\natexlab{c}}.
\newblock Improved baselines with momentum contrastive learning.
\newblock \emph{arXiv preprint arXiv:2003.04297}.

\bibitem[{Cubuk et~al.(2020)Cubuk, Zoph, Shlens, and Le}]{cubuk2020randaugment}
Cubuk, E.~D.; Zoph, B.; Shlens, J.; and Le, Q.~V. 2020.
\newblock Randaugment: Practical automated data augmentation with a reduced
  search space.
\newblock In \emph{Proceedings of the IEEE/CVF Conference on Computer Vision
  and Pattern Recognition Workshops}, 702--703.

\bibitem[{Ghosh and Thiery(2021)}]{ghosh2021data}
Ghosh, A.; and Thiery, A.~H. 2021.
\newblock On Data-Augmentation and Consistency-Based Semi-Supervised Learning.
\newblock In \emph{International Conference on Learning Representations}.

\bibitem[{Grandvalet, Bengio et~al.(2005)}]{grandvalet2005semi}
Grandvalet, Y.; Bengio, Y.; et~al. 2005.
\newblock Semi-supervised learning by entropy minimization.
\newblock In \emph{CAP}, 281--296.

\bibitem[{He et~al.(2020)He, Fan, Wu, Xie, and Girshick}]{he2020momentum}
He, K.; Fan, H.; Wu, Y.; Xie, S.; and Girshick, R. 2020.
\newblock Momentum contrast for unsupervised visual representation learning.
\newblock In \emph{Proceedings of the IEEE/CVF Conference on Computer Vision
  and Pattern Recognition}, 9729--9738.

\bibitem[{Iscen et~al.(2019)Iscen, Tolias, Avrithis, and Chum}]{iscen2019label}
Iscen, A.; Tolias, G.; Avrithis, Y.; and Chum, O. 2019.
\newblock Label propagation for deep semi-supervised learning.
\newblock In \emph{Proceedings of the IEEE/CVF Conference on Computer Vision
  and Pattern Recognition}, 5070--5079.

\bibitem[{Jawed, Grabocka, and Schmidt-Thieme(2020)}]{jawed2020self}
Jawed, S.; Grabocka, J.; and Schmidt-Thieme, L. 2020.
\newblock Self-supervised learning for semi-supervised time series
  classification.
\newblock In \emph{Pacific-Asia Conference on Knowledge Discovery and Data
  Mining}, 499--511. Springer.

\bibitem[{Khosla et~al.(2020)Khosla, Teterwak, Wang, Sarna, Tian, Isola,
  Maschinot, Liu, and Krishnan}]{khosla2020supervised}
Khosla, P.; Teterwak, P.; Wang, C.; Sarna, A.; Tian, Y.; Isola, P.; Maschinot,
  A.; Liu, C.; and Krishnan, D. 2020.
\newblock Supervised Contrastive Learning.
\newblock \emph{Advances in Neural Information Processing Systems}, 33.

\bibitem[{Kim et~al.(2021)Kim, Choo, Kwon, Joe, Min, and
  Gwon}]{kim2021selfmatch}
Kim, B.; Choo, J.; Kwon, Y.-D.; Joe, S.; Min, S.; and Gwon, Y. 2021.
\newblock SelfMatch: Combining Contrastive Self-Supervision and Consistency for
  Semi-Supervised Learning.
\newblock \emph{arXiv preprint arXiv:2101.06480}.

\bibitem[{Kolesnikov et~al.(2020)Kolesnikov, Beyer, Zhai, Puigcerver, Yung,
  Gelly, and Houlsby}]{kolesnikov2019large}
Kolesnikov, A.; Beyer, L.; Zhai, X.; Puigcerver, J.; Yung, J.; Gelly, S.; and
  Houlsby, N. 2020.
\newblock Big transfer (bit): General visual representation learning.
\newblock In \emph{Computer Vision--ECCV 2020: 16th European Conference,
  Glasgow, UK, August 23--28, 2020, Proceedings, Part V 16}, 491--507.
  Springer.

\bibitem[{Lee et~al.(2013)}]{lee2013pseudo}
Lee, D.-H.; et~al. 2013.
\newblock Pseudo-label: The simple and efficient semi-supervised learning
  method for deep neural networks.
\newblock In \emph{Workshop on challenges in representation learning, ICML},
  volume~3.

\bibitem[{Li, Xiong, and Hoi(2020)}]{li2020comatch}
Li, J.; Xiong, C.; and Hoi, S. 2020.
\newblock CoMatch: Semi-supervised Learning with Contrastive Graph
  Regularization.
\newblock \emph{arXiv preprint arXiv:2011.11183}.

\bibitem[{Miyato et~al.(2018)Miyato, Maeda, Koyama, and
  Ishii}]{miyato2018virtual}
Miyato, T.; Maeda, S.-i.; Koyama, M.; and Ishii, S. 2018.
\newblock Virtual adversarial training: a regularization method for supervised
  and semi-supervised learning.
\newblock \emph{IEEE transactions on pattern analysis and machine
  intelligence}, 41(8): 1979--1993.

\bibitem[{Oliver et~al.(2018)Oliver, Odena, Raffel, Cubuk, and
  Goodfellow}]{oliver2018realistic}
Oliver, A.; Odena, A.; Raffel, C.~A.; Cubuk, E.~D.; and Goodfellow, I. 2018.
\newblock Realistic Evaluation of Deep Semi-Supervised Learning Algorithms.
\newblock \emph{Advances in Neural Information Processing Systems}, 31:
  3235--3246.

\bibitem[{Paszke et~al.(2019)Paszke, Gross, Massa, Lerer, Bradbury, Chanan,
  Killeen, Lin, Gimelshein, Antiga, Desmaison, Kopf, Yang, DeVito, Raison,
  Tejani, Chilamkurthy, Steiner, Fang, Bai, and Chintala}]{NEURIPS2019_9015}
Paszke, A.; Gross, S.; Massa, F.; Lerer, A.; Bradbury, J.; Chanan, G.; Killeen,
  T.; Lin, Z.; Gimelshein, N.; Antiga, L.; Desmaison, A.; Kopf, A.; Yang, E.;
  DeVito, Z.; Raison, M.; Tejani, A.; Chilamkurthy, S.; Steiner, B.; Fang, L.;
  Bai, J.; and Chintala, S. 2019.
\newblock PyTorch: An Imperative Style, High-Performance Deep Learning Library.
\newblock In Wallach, H.; Larochelle, H.; Beygelzimer, A.; d\textquotesingle
  Alch\'{e}-Buc, F.; Fox, E.; and Garnett, R., eds., \emph{Advances in Neural
  Information Processing Systems 32}, 8024--8035. Curran Associates, Inc.

\bibitem[{Rousseeuw(1987)}]{rousseeuw1987silhouettes}
Rousseeuw, P.~J. 1987.
\newblock Silhouettes: a graphical aid to the interpretation and validation of
  cluster analysis.
\newblock \emph{Journal of computational and applied mathematics}, 20: 53--65.

\bibitem[{Sajjadi, Javanmardi, and Tasdizen(2016)}]{sajjadi2016regularization}
Sajjadi, M.; Javanmardi, M.; and Tasdizen, T. 2016.
\newblock Regularization with stochastic transformations and perturbations for
  deep semi-supervised learning.
\newblock \emph{Advances in neural information processing systems}, 29:
  1163--1171.

\bibitem[{Samuli and Timo(2017)}]{laine2016temporal}
Samuli, L.; and Timo, A. 2017.
\newblock Temporal ensembling for semi-supervised learning.
\newblock In \emph{International Conference on Learning Representations}.

\bibitem[{Shi et~al.(2018)Shi, Gong, Ding, Tao, and
  Zheng}]{shi2018transductive}
Shi, W.; Gong, Y.; Ding, C.; Tao, Z.~M.; and Zheng, N. 2018.
\newblock Transductive semi-supervised deep learning using min-max features.
\newblock In \emph{Proceedings of the European Conference on Computer Vision
  (ECCV)}, 299--315.

\bibitem[{Sohn et~al.(2020)Sohn, Berthelot, Carlini, Zhang, Zhang, Raffel,
  Cubuk, Kurakin, and Li}]{sohn2020fixmatch}
Sohn, K.; Berthelot, D.; Carlini, N.; Zhang, Z.; Zhang, H.; Raffel, C.~A.;
  Cubuk, E.~D.; Kurakin, A.; and Li, C.-L. 2020.
\newblock FixMatch: Simplifying Semi-Supervised Learning with Consistency and
  Confidence.
\newblock \emph{Advances in Neural Information Processing Systems}, 33.

\bibitem[{Tarvainen and Valpola(2017)}]{tarvainen2017mean}
Tarvainen, A.; and Valpola, H. 2017.
\newblock Mean teachers are better role models: Weight-averaged consistency
  targets improve semi-supervised deep learning results.
\newblock \emph{Advances in Neural Information Processing Systems}, 30.

\bibitem[{Verma et~al.(2019)Verma, Lamb, Kannala, Bengio, and
  Lopez-Paz}]{verma2019interpolation}
Verma, V.; Lamb, A.; Kannala, J.; Bengio, Y.; and Lopez-Paz, D. 2019.
\newblock Interpolation Consistency Training for Semi-supervised Learning.
\newblock In \emph{Proceedings of the 28th International Joint Conference on
  Artificial Intelligence}, IJCAI'19, 3635--3641. AAAI Press.
\newblock ISBN 978-0-9992411-4-1.

\bibitem[{Vincent et~al.(2010)Vincent, Larochelle, Lajoie, Bengio, Manzagol,
  and Bottou}]{vincent2010stacked}
Vincent, P.; Larochelle, H.; Lajoie, I.; Bengio, Y.; Manzagol, P.-A.; and
  Bottou, L. 2010.
\newblock Stacked denoising autoencoders: Learning useful representations in a
  deep network with a local denoising criterion.
\newblock \emph{Journal of machine learning research}, 11(12).

\bibitem[{Xie et~al.(2020)Xie, Dai, Hovy, Luong, and Le}]{xie2020unsupervised}
Xie, Q.; Dai, Z.; Hovy, E.; Luong, T.; and Le, Q. 2020.
\newblock Unsupervised Data Augmentation for Consistency Training.
\newblock \emph{Advances in Neural Information Processing Systems}, 33.

\bibitem[{Yu et~al.(2020)Yu, Ikami, Irie, and Aizawa}]{yu2020multi}
Yu, Q.; Ikami, D.; Irie, G.; and Aizawa, K. 2020.
\newblock Multi-Task Curriculum Framework for Open-Set Semi-Supervised
  Learning.
\newblock In \emph{European Conference on Computer Vision}, 438--454. Springer.

\bibitem[{Zagoruyko and Komodakis(2016)}]{zagoruyko2016wide}
Zagoruyko, S.; and Komodakis, N. 2016.
\newblock Wide Residual Networks.
\newblock In \emph{British Machine Vision Conference 2016}. British Machine
  Vision Association.

\bibitem[{Zamir et~al.(2018)Zamir, Sax, Shen, Guibas, Malik, and
  Savarese}]{zamir2018taskonomy}
Zamir, A.~R.; Sax, A.; Shen, W.; Guibas, L.~J.; Malik, J.; and Savarese, S.
  2018.
\newblock Taskonomy: Disentangling task transfer learning.
\newblock In \emph{Proceedings of the IEEE conference on computer vision and
  pattern recognition}, 3712--3722.

\bibitem[{Zhai et~al.(2019)Zhai, Oliver, Kolesnikov, and Beyer}]{zhai2019s4l}
Zhai, X.; Oliver, A.; Kolesnikov, A.; and Beyer, L. 2019.
\newblock S4l: Self-supervised semi-supervised learning.
\newblock In \emph{Proceedings of the IEEE/CVF International Conference on
  Computer Vision}, 1476--1485.

\bibitem[{Zhang et~al.(2018)Zhang, Cisse, Dauphin, and
  Lopez-Paz}]{zhang2018mixup}
Zhang, H.; Cisse, M.; Dauphin, Y.~N.; and Lopez-Paz, D. 2018.
\newblock mixup: Beyond Empirical Risk Minimization.
\newblock In \emph{International Conference on Learning Representations}.

\end{thebibliography}
}

\clearpage
\appendix
\onecolumn

\section{A. Contrastive Regularization for Semi-Supervised Learning}
In this section, we introduce the detailed derivation of the gradients of contrastive regularization.
For a model parameterized by $\theta$, the label prediction of an unlabeled sample $u \in \mathcal{U}$ is $\hat{p}(y|u)=\text{softmax}[W^\top h_\theta (u)]$, where a mini-batch of unlabeled sample $\mathcal{U}$, $K$-class weight matrix  $W=[w_1, w_2, ..., w_K] \in \mathbb{R}^{H \times K}$ and the penultimate features of $u$ $h_\theta (u) \in \mathbb{R}^{H}$.
We assume that the model parameter $\theta$ comprises the class weight matrix $W$.
For a stochastic strong augmentation $\alpha$, we define the set of $m$ strongly augmented samples in an unlabeled mini-batch as $\mathcal{A}_m(\mathcal{U})=\{u'_{i} | u \in \mathcal{U}, u'_{i} = \alpha(u), 1 \leq i \leq m \}$.
$\hat{q}_{u}$ is the pseudo-label of $u'$ and defined as $\hat{q}_{u}= \argmax q_{u}$, where $q_{u}= {\text{sg}[\hat{p}(y|u)}]$ and sg is the stop gradient.
We define the set of \textit{pseudo}-positive pairs of $u'$ as $\hat{P}(u')=\{ p' | p' \in \mathcal{A}_m(\mathcal{U})/u', \hat{q}_p = \hat{q}_u \}$, where $\hat{q}_p$ and $\hat{q}_u$ are the pseudo-label of $p'$ and $u'$, respectively.
Note that pseudo-labels of strongly augmented samples are defined by the label predictions on the samples \textit{before} the strong data augmentation to improve the reliability of pseudo-labeling.
Then, for an unlabeled sample $u$ in an unlabeled mini-batch $\mathcal{U}$, contrastive regularization, $\mathcal{R}_{CR}$, is defined as follows:
\begin{equation}
\mathcal{R}_{CR} (\mathcal{U}) = \frac{1}{|\mathcal{A}_m(\mathcal{U})|}\sum_{u'\in \mathcal{A}_m(\mathcal{U})} \mathds{1}[\max q_{u} > \delta'] r(u'), 
\label{eq:batch_CR}    
\end{equation}
\begin{equation}
r (u') = \frac{-1}{|\hat{P}(u')|} \sum_{p' \in \hat{P}(u')} \log \frac{\exp (\langle z_{u'}, z_{p'} \rangle/ \tau )}{\sum_{v' \in \mathcal{A}_m(\mathcal{U})/u'} \exp (\langle z_{u'}, z_{v'} \rangle / \tau )}, \label{eq:sample_CR}
\end{equation}
where a confidence threshold $\delta'$, a temperature scaling parameter$\tau$, and a normalized vector of the projection head output $z_{u'}$.

Assuming that an augmented sample $u'=\alpha{A}(u)$ has a confident pseudo-label, we first show that the contrastive regularization moves the features of $u'$ toward the centroid of the feature cluster having the same pseudo-label, while pushing away the features in different clusters.
Without the loss of generalizability, we assume $z=h$ and $\tau=1$.
In addition, we omit $\theta$ for the notation brevity.
The first-order derivative of $\mathcal{R}_{CR}$ with respect to a feature vector of $u'$ is as follows:
\begin{align}
    -\frac{\partial r(u')}{\partial h(u')} &= \frac{-1}{|\hat{P}(u')|}\sum_{p'\in \hat{P}(u')} ( -h(p') + \sum_{v' \in \mathcal{A}(\mathcal{U})/u'} s[u',v']h(v') ) \label{eq:supp_feature_mean}\\
                                            &= \frac{-1}{|\hat{P}(u')|}\sum_{p'\in \hat{P}(u')} [ - h(p') + \sum_{v'_p \in \hat{P}(u')}s[u', v'_p]h(v'_p) + \sum_{v'_n \in N(u')}s[u', v'_n]h(v'_n) ]\\
                                            &= \sum_{p'\in \hat{P}(u')} (\frac{1}{|\hat{P}(u')|} - s[u', p']) h(p') - \sum_{v'_n \in N(u')}s[u', v'_n]h(v'_n) \label{eq:supp_pos_neg}
\end{align}
where $s[u',p']$ is the softmax score of the pair of $h(u')$ and $h(p')$, $N(u')=\{n'|n' \in \mathcal{A}_m(\mathcal{U}), \hat{q}_n \neq \hat{q}_u\}$, $n$ is the original sample of strongly augmented $n'$.
Since the sum of minus log-sum-exp terms is the convex function, the first-order optimality condition holds when the $s^{*}[u',v']=1/|\hat{P}(u')|$ if $v' \in \hat{P}(u')$ and $s^{*}[u',v']=0$ otherwise.
Thus, the contrastive regularization on an unlabeled sample pushes the features of different pseudo-labels and pulls those of the same pseudo-label.
Assuming that the softmax scores of the negative pairs are small enough, Eq.~(\ref{eq:supp_pos_neg}) is summarized as follows:
\begin{equation}
    -\frac{\partial r(u')}{\partial h(u')} = \sum_{p'\in \hat{P}(u')} (\frac{1}{|\hat{P}(u')|} - s[u', p']) h(p') + R(u'),
\end{equation}
where $R(u')$ is a remainder term.
The feature vector of $u'$ is updated toward the centroid, which is the weighted sum of positive features regardless of the confidence of the pseudo-labels.

For another unlabeled sample $v\in\mathcal{U}$ and $v'=\mathcal{A}(v)$, the contrastive regularization on features with confident pseudo-labels pushes the features of $v'$ if $v'$ has a different pseudo-label, and pulls them otherwise.
By the same process of the derivation of above $-\partial r(u') / \partial h(u')$, we can derive $-\partial r(u') / \partial h(v')$ as follows:
\begin{equation} \label{eq:supp_gradient_cr_unconf} 
    -\frac{\partial r(u')}{\partial h_\theta(v')} = 
    \begin{cases}
     (\frac{1}{|\hat{P}(u')|} - s[u',v'])h_\theta(u'), &  \mbox{if } v' \in \hat{P}(u') \\
     ~~~~ - s[u',v']h_\theta(u'), & \mbox{if } v' \notin \hat{P}(u')
    \end{cases}.
\end{equation}

\section{B. Implementation Details}
For the implementation, we follow the setting of the original FixMatch paper \cite{sohn2020fixmatch} except the hyper-parameters related to the contrastive regularization.
We use Pytorch 1.6.0 to reproduce FixMatch and implement the contrastive regularization on the same codebase.
We conduct all experiments using four Tesla V100 32GB GPUs, except the ImageNet dataset.
For ImageNet, we use 32 V100 GPUs for FixMatch and 64 GPUs for FixMatch with the contrastive regularization.
For all datasets, we use the stochastic gradient descent (SGD) optimizer with Nesterov momentum $\beta=0.9$, and temperature parameter $\tau = 0.01$.
We use an exponential moving average (EMA) of model parameters with 0.999 momentum and cosine learning scheduling used in \cite{sohn2020fixmatch}.
The batch size of labeled data ($B$) is 64 for SVHN, CIFAR-10, CIFAR-100, and STL-10 except SVHN with 20 and 40 labels, and CIFAR-10 with 20 labels.
Considering the small number of labeled samples, we use 16 labeled samples for training SVHN with 20 and 40 labels, and 32 labeled samples for CIFAR-10 with 20 labels per training iteration of WRN-28-2.
For WRN-28-8, we use 16 labeled samples for training SVHN with 20 and 40 labels, and 32 labeled samples for CIFAR-10 with 20 and 40 labels.

In the ablation study for different hyper-parameters, we use WRN-28-4 for CIFAR-100 with 2500 labels, because WRN-28-8 requires over two times more training time than WRN-28-4 but the accuracy gain is marginal (+0.83\% in Table~\ref{tab:abl_wf}).
We use random horizontal flipping and the random crop for both weak and strong augmentations of training datasets.
For the SVHN dataset, we do not use horizontal flipping, considering that a classifier can be easily confused to discriminate some classes such as eight and three.
For strong augmentations, we use RandAugment \cite{sohn2020fixmatch} following the original paper of FixMatch.
Please refer to the original paper of FixMatch and our source codes to check the details of augmentation policies according to the datasets.
The other training details are available in Table~\ref{tab:abl_setting}.

\begin{table}[]
\centering
\small
\caption{The details of hyper-parameters on training datasets. LR and WD describe initial learning rate and weight decay, respectively.}
\begin{tabular}{c|cccccccccc} \toprule
Dataset   & Model     & $B$  & $\mu$ & Epochs & $\lambda_\text{CS}$ & $\lambda_\text{CR}$ & LR & WD & $\delta$ & $\delta'$ \\ \hline
SVHN      & WRN-28-2  & 16, 64   & 7     & 6500 & 1.0                 & 1.0                 & 0.03          & 0.0005       & 0.95     & 0.95      \\
CIFAR-10  & WRN-28-2  & 32, 64   & 7  & 6500   & 1.0                 & 1.0                 & 0.03          & 0.0005       & 0.95     & 0.95      \\
CIFAR-100 & WRN-28-8  & 64   & 7    & 2500  & 1.0                 & 10.0                & 0.03          & 0.001        & 0.95     & 0.95      \\
STL-10    & WRN-37-2  & 64   & 7 & 5000     & 1.0                 & 10.0                & 0.03          & 0.0005       & 0.95     & 0.95      \\
ImageNet  & ResNet-50 & 1024 & 5  & 300   & 10.0                & 1.0                 & 0.05          & 0.003       & 0.7      & 0.7      \\ \bottomrule
\end{tabular}
\label{tab:abl_setting}
\end{table}

\section{C. Additional Experimental Results}

\begin{wrapfigure}{r} {0.3\textwidth}
\centering
\includegraphics[width=0.3\textwidth]{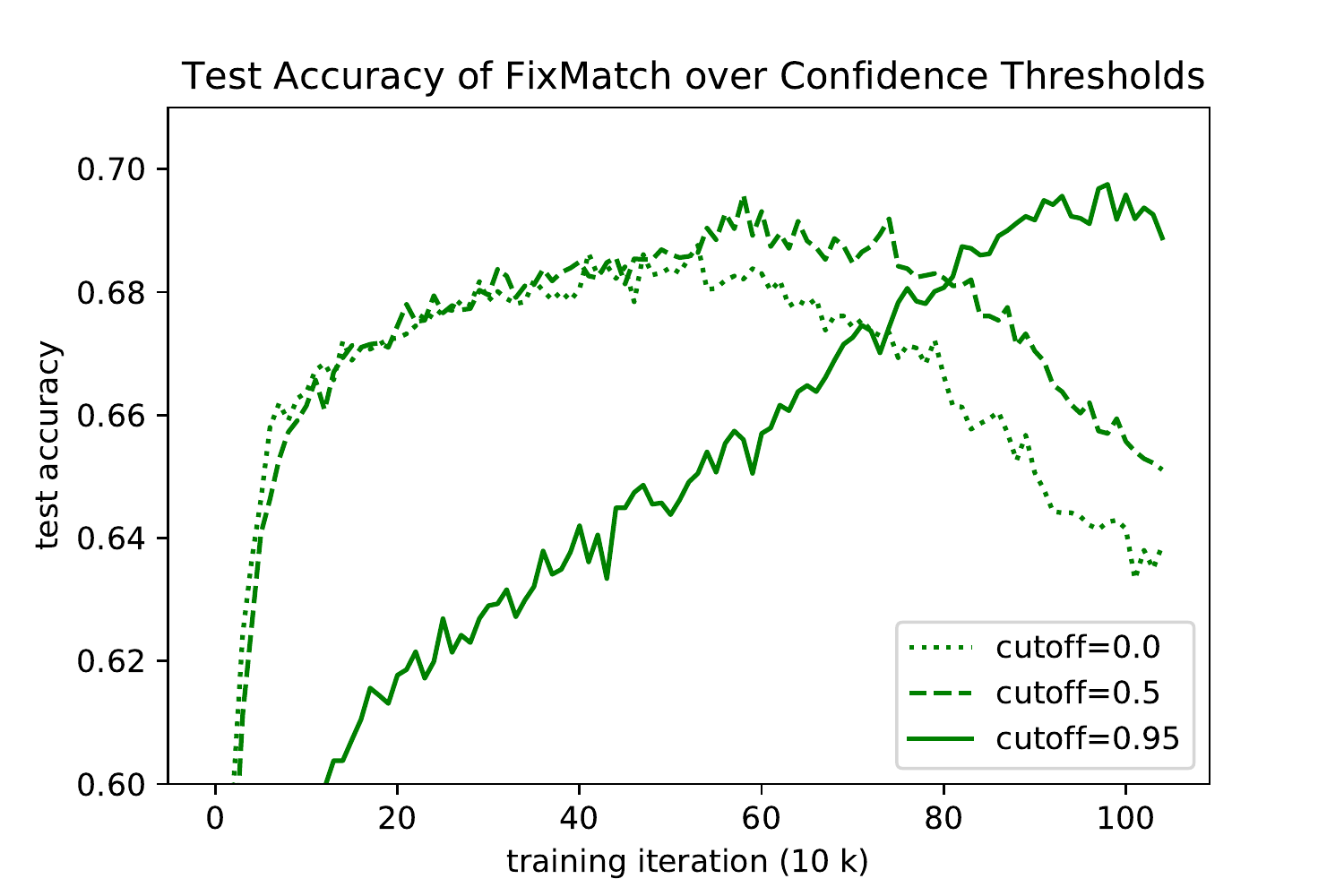}
\caption{Test accuracy of FixMatch with different confident thresholds views over training.}
\label{fig:supp_thres}
\end{wrapfigure}

In this study, we have claimed that the previous consistency regularization suffers from the training inefficiency by the exclusion of unconfident pseudo-labels to ensure the reliability of pseudo-labeling.
Figure~\ref{fig:supp_thres} shows the relationship between the confident threshold $\delta$ and the convergence speed of SSL training in FixMatch.
When low confidence thresholds are used such as 0.0 or 0.7, FixMatch can leverage more unlabeled samples than FixMatch with $\delta=0.95$.
Thus, SSL training with low confidence thresholds can achieve the best performance much faster and increase the training efficiency with respect to the training iterations and time.
However, the low $\delta$s lower the best test accuracies, because the unreliable pseudo-labeling propagates wrong labeling information to other unlabeled samples.
Meanwhile, when a high confidence threshold is used, the accuracy increases slower especially in the early stage of training, because it cannot leverage many unlabeled samples.
Despite the low training efficiency, FixMatch with $\delta=0.95$ can achieve high performance, while leveraging only its confident labeling information.
The results imply that the training inefficiency of the previous consistency regularization comes from the trade-off between the reliability of pseudo-labeling and the number of used unlabeled samples in training.

\begin{wrapfigure}{r} {0.3\textwidth}
\centering
\includegraphics[width=0.3\textwidth]{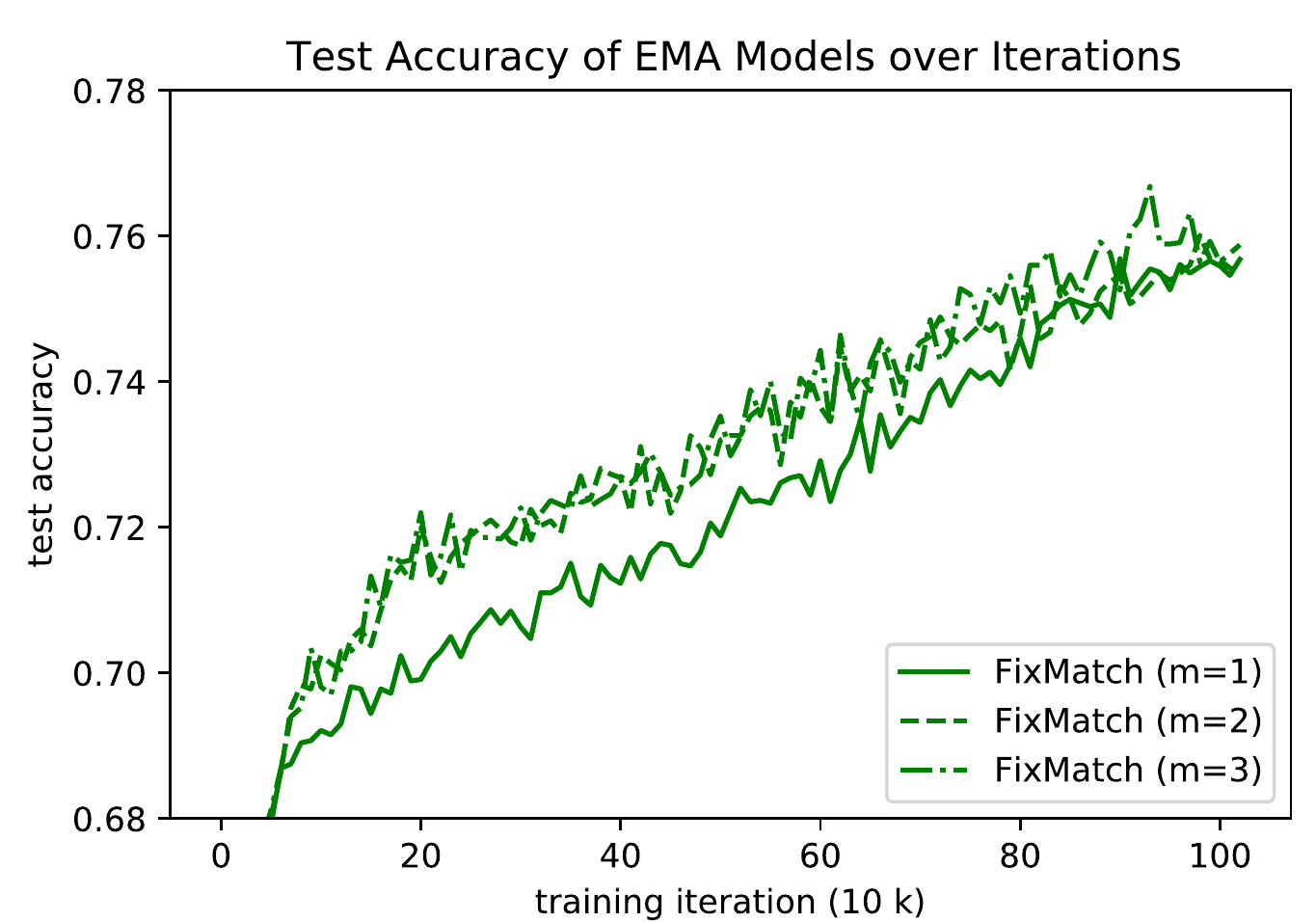}
\caption{Test accuracy of FixMatch with different views over training.}
\label{fig:supp_view}
\end{wrapfigure}

In the ablation study, we have shown that the increased views ($m$) of unlabeled samples also increase the accuracy of FixMatch.
However, we show that more numbers of views cannot improve the training efficiency, and the accuracy over training still increases gradually in the training.
The results imply that the effect of our contrastive regularization does not result from the increased views of unlabeled samples, but well-clustered representations for SSL.

\begin{figure}
\centering
\includegraphics[width=0.6\textwidth]{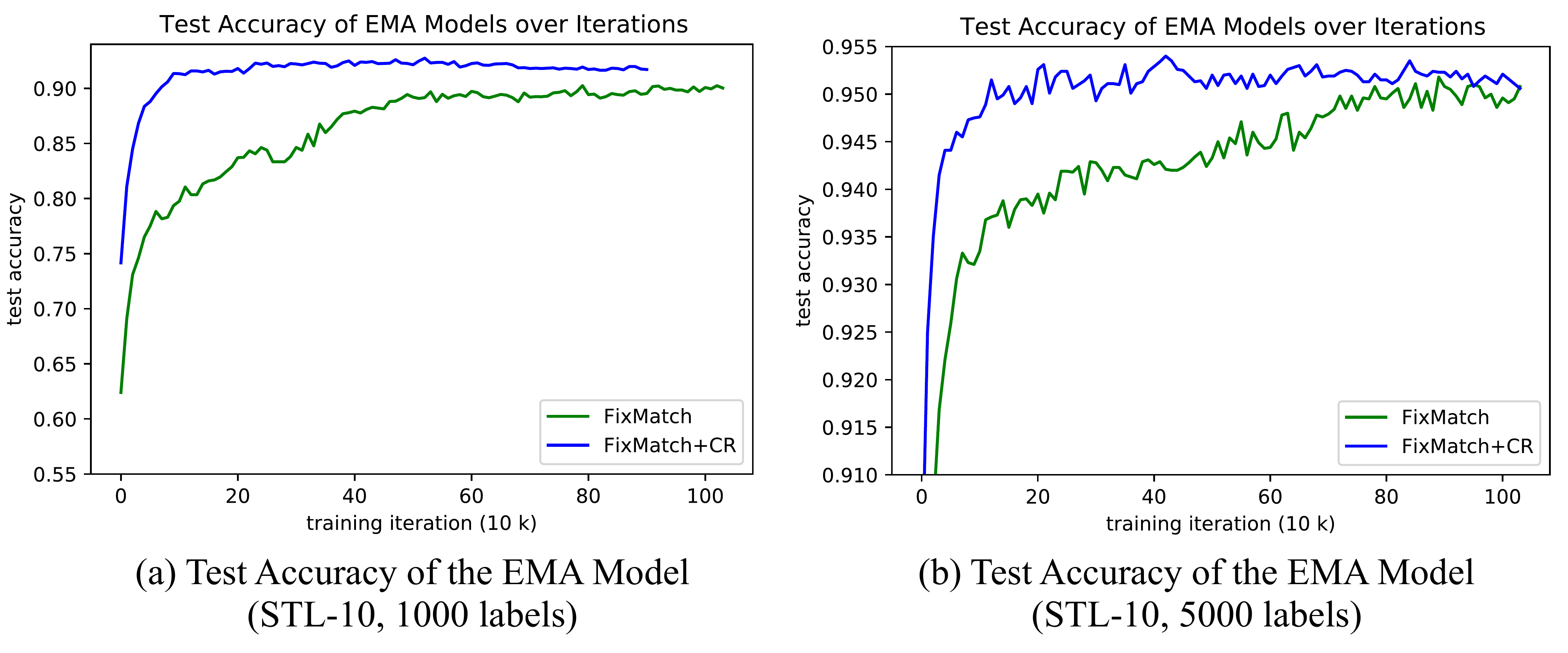}
\caption{Test accuracy of FixMatch and FixMatch+CR of WRN-37-2 trained on STL-10 with (a) 1000 and (b) 5000 labels.}
\label{fig:supp_stl}
\end{figure}

\begin{figure}
\centering
\includegraphics[width=\textwidth]{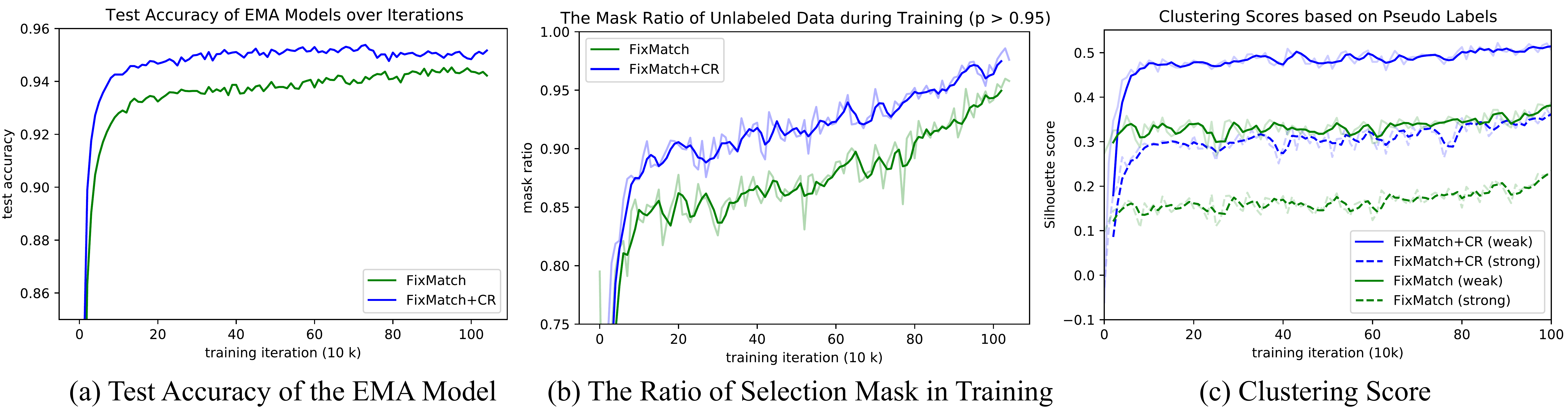}
\caption{Empirical Results of FixMatch and FixMatch+CR with WRN-28-2 trained on the CIFAR-10 with 250 labels. (a) Test accuracy of EMA models, (b) the ratio of selection mask, and (c) Silhouette score of penultimate features of unlabeled samples based on pseudo-labels.}
\label{fig:supp_c10_250}
\end{figure}

\begin{figure}
\centering
\includegraphics[width=\textwidth]{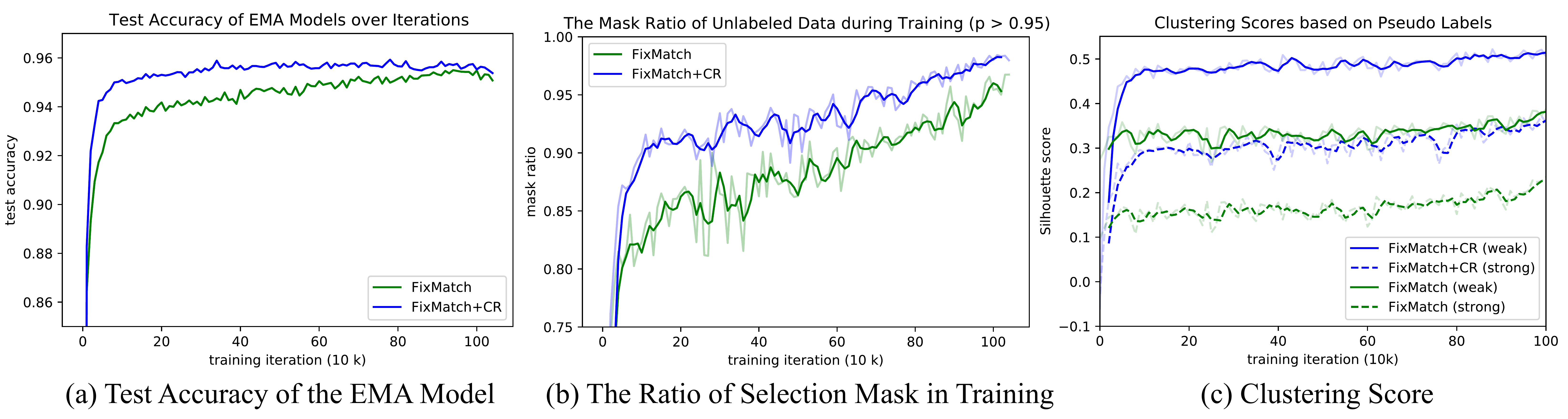}
\caption{Empirical Results of FixMatch and FixMatch+CR with WRN-28-2 trained on the CIFAR-10 with 4000 labels. (a) Test accuracy of EMA models, (b) the ratio of selection mask, and (c) Silhouette score of penultimate features of unlabeled samples based on pseudo-labels.}
\label{fig:supp_c10_4000}
\end{figure}

\begin{figure}
\centering
\includegraphics[width=\textwidth]{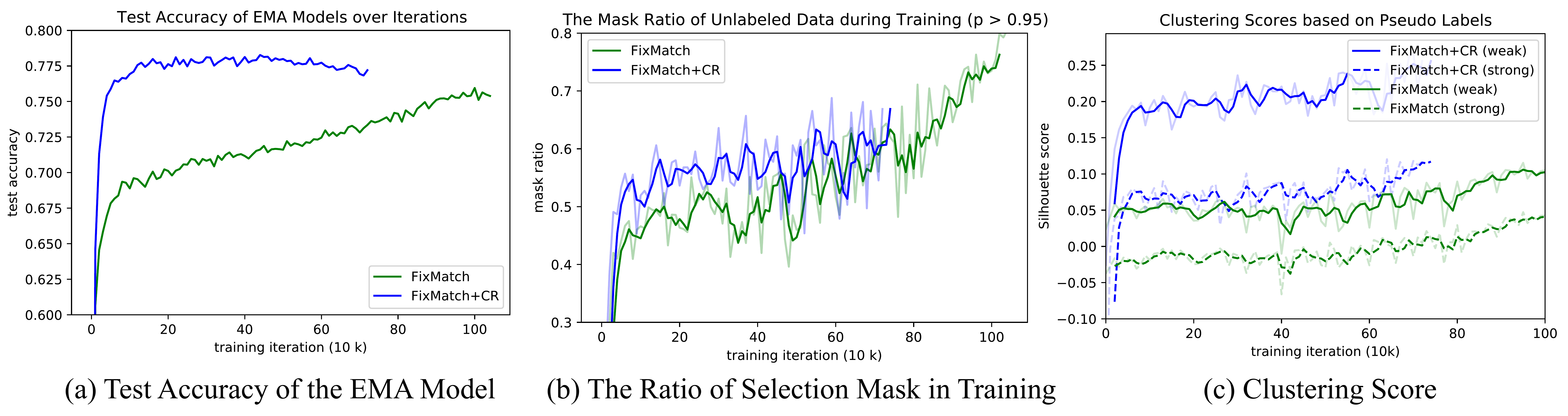}
\caption{Empirical Results of FixMatch and FixMatch+CR with WRN-28-4 trained on the CIFAR-100 with 1000 labels. (a) Test accuracy of EMA models, (b) the ratio of selection mask, and (c) Silhouette score of penultimate features of unlabeled samples based on pseudo-labels.}
\label{fig:supp_c100}
\end{figure}

We show the additional results to show the more efficient training of contrastive regularization than consistency regularization.
For better visualization, we try to train FixMatch+CR in the same epochs of FixMatch, 10,500 epochs, until FixMatch+CR starts to be overfitted.
First, we show that the results described in Section 4.3 are also consistent with other datsets such as CIFAR-10 (Figure~\ref{fig:supp_c100}) and STL-10 (Figure~\ref{fig:supp_stl}).
Figure~\ref{fig:supp_c10_250} and ~\ref{fig:supp_c10_4000} show the test accuracy of EMA models, the average ratio of selection mask in a training mini-batch, and clustering scores of features in training on the CIFAR-10 dataset.
The results imply that our contrastive regularization learns well-clustered features for SSL and requires fewer iterations for high performance.
The accuracy of FixMatch gradually increases over the entire training iterations, but the accuracy of FixMatch+CR increases much faster than FixMatch, especially in the early stage of training.
Moreover, we find that about 10\% of iterations for FixMatch+CR are enough to achieve the performance of FixMatch.


\end{document}